\pdfoutput=1
\documentclass[11pt]{article}

\usepackage[preprint]{acl}
\usepackage{placeins}

\usepackage{times}
\usepackage{latexsym} 
\usepackage{booktabs}
\usepackage{subfiles}
\usepackage{graphicx}
\usepackage{array}
\usepackage{amsmath}   % for \mathbb{E}, \mathcal{L}, aligned math
\usepackage{amssymb}
\usepackage{amsmath}
\usepackage{tabularx}
\usepackage{float}
\usepackage{graphicx, makecell}
\usepackage[T1]{fontenc}
\usepackage{calc}
\usepackage[utf8]{inputenc}

\usepackage{microtype}
\usepackage{subfiles}
\usepackage{inconsolata}

\usepackage{graphicx}
\usepackage{pifont}
\usepackage{xcolor}

\usepackage[most]{tcolorbox} % loads many useful tcolorbox libraries
\usepackage{xcolor}

\newtcolorbox{promptbox}[1]{%
  enhanced,
  colback=blue!5,
  colframe=blue!60!black,
  coltitle=white,
  fonttitle=\bfseries,
  title=#1, % <-- dynamic title
  attach boxed title to top left={xshift=6pt,yshift=-2pt},
  boxed title style={
    colback=blue!60!black,
    colframe=blue!60!black,
    arc=4pt,
    left=6pt,right=6pt,top=3pt,bottom=3pt
  },
  arc=6pt,
  boxrule=0.8pt,
  left=10pt,right=10pt,top=8pt,bottom=8pt,
  drop shadow
}

\title{Few-Shot Large Language Models for Actionable Triage Categorization of Online Patient Inquiries}

\author{
    Liqi Zhou \\
    \texttt{zliqi@seas.upenn.edu}
    \And
    Jiafu Li \\
    \texttt{ericlijiafu@alum.northwestern.edu}
}

\begin{document}
\maketitle
\begin{abstract}
Online patient inquiries are often informal, incomplete, and written before professional assessment, yet they must still be routed to an appropriate level of clinical follow-up. We study this as a four-class actionable triage task -- self-care, schedule-visit, urgent-clinician-review, or emergency-referral, and ask whether prompted large language models (LLMs) can support such routing under low-resource labeling conditions. Using the public HealthCareMagic-100K corpus, we construct a 300-example human calibrated gold evaluation set, a 700-example auto-labeled silver training set, and a 40-example few-shot pool. We compare Term Frequency-Inverse Document Frequency (TF-IDF) and Bidirectional Encoder Representations from Transformers for Biomedical Text Mining (BioBERT) baselines train on silver labels against six prompted LLMs under 0-shot, 4-shot, and 12-shot conditions respectively. Accordingly, we evaluate with macro-$F_1$ alongside safety-aware metrics, including emergency-recall, under-triage rate, and severe under-triage rate. The strongest LLM (Claude Haiku 4.5, 12-shot) reaches macro-$F_1$ 0.475, exceeding the best supervised baseline (BioBERT, 0.378) on point estimate, with overlapping confidence intervals. Few-shot prompting and two-model agreement help in label-dependent ways: self-care agreement is reliable, urgent-clinician-review is not. We conclude that LLMs can support triage prioritization and selective human review, but not autonomous deployment.

\end{abstract}

\section{Introduction}

Online patient inquiries contain unstructured symptom descriptions, medication questions, test result concerns, and care-seeking requests. These messages are often written before professional assessment and may be incomplete, informal, or clinically ambiguous. Health systems and online care platforms, therefore, need scalable methods to help route patient-authored messages to an appropriate level of follow-up.

This study focuses on actionable triage categorization rather than diagnosis or treatment recommendations. Given a patient inquiry, the task is to assign exactly one of four care-routing labels: self-care, schedule-visit, urgent-clinician-review, or emergency-referral. This framing emphasizes the operational question of what type of response is appropriate, not the clinical question of what condition the patient has.

Automatic triage is challenging for three reasons. First, patient-authored text often lacks key clinical details such as duration, severity, comorbidities, vital signs, or current care status. Second, high-acuity cases are relatively rare but safety critical. Third, standard aggregate metrics such as accuracy or even macro-$F_1$ may obscure clinically asymmetric errors. In triage, under-triage of urgent or emergency cases is typically more concerning than conservative over-triage, so evaluation should explicitly measure high-acuity recall, false negatives, and severity-direction errors.

Large language models (LLMs) may be useful for this setting because they can interpret free-text symptom descriptions, follow label definitions, and adapt through prompting without task-specific fine-tuning. However, their value for online patient inquiry triage cannot be assumed. Prior work has studied clinical text classification, patient-message urgency, emergency department acuity assessment, and LLM-based medical natural language processing (NLP), but less is known about how few-shot LLMs behave in a low-resource, action-oriented triage setting with human-calibrated labels and safety-aware evaluation. In particular, it remains unclear whether prompted LLMs offer practical advantages over lightweight supervised baselines, how sensitive they are to prompt design, and whether model agreement can support selective human review.

We address these questions using public online medical dialogue data and a guideline-informed, human calibrated gold evaluation set. We compare supervised Term Frequency-Inverse Document Frequency (TF-IDF) and Bidirectional Encoder Representations from Transformers for Biomedical Text Mining (BioBERT) baselines against open-source and API-served LLMs under zero-shot and few-shot prompting conditions. We evaluate models using both standard classification metrics and safety-aware triage metrics; furthermore, we analyze model disagreement as a signal for human-in-the-loop (HITL) selective review. We summarize our
contributions as follows:

\begin{itemize}

    \item
      We formulate online patient inquiry triage as a four-class actionable routing task rather than as diagnosis generation or general medical text classification.
    \item
      We construct a guideline-informed, human-calibrated gold evaluation set from public medical dialogue data, with separate silver training and few-shot prompt-construction splits.
    \item
      We compare supervised TF-IDF and BioBERT baselines against multiple prompted LLMs under 0-shot, 4-shot, and 12-shot settings.
    \item
      We evaluate triage performance using safety-aware metrics, including urgent and emergency recall, under-triage, severe under-triage, and high-acuity false negatives.
    \item
      We explore two-model consensus as an oracle HITL selective prediction strategy, showing that model agreement is label-dependent and should not be treated as a universal auto-accept signal.
      
\end{itemize}
\section{Related Work}

\subsection{Clinical Text Classification and Triage NLP}\label{clinical-text-classification-and-triage-nlp}

Clinical text classification has long supported triage and routing tasks. In emergency departments, machine learning and clinical NLP have been applied to predict acuity from triage documentation and clinical notes \cite{Ivanov2021}, and broader reviews show sustained interest while  emphasizing heterogeneity in clinical settings, data sources, and evaluation designs \cite{Porto2024}. A separate line of work moves from clinician-authored documentation to patient-authored messages: \citet{Gatto2022} study perceived severity in patient-generated telemedical COVID queries, and \citet{Si2020} evaluate BERT-based methods for patient-message triage in small-data settings. These studies are directly relevant because patient-authored text is informal, incomplete, and written before professional assessment, but they remain primarily supervised task-specific systems and do not address safety-aware, action oriented routing over public online medical dialogue.

\subsection{LLMs in Healthcare NLP}\label{large-language-models-in-healthcare-nlp}

LLMs have made zero-shot and few-shot prompting a practical alternative to task-specific supervised training in healthcare NLP. However, prior evidence does not support assuming that prompted LLMs are uniformly superior. Large-scale biomedical NLP benchmarking shows substantial variation by task and evaluation design \cite{Chen2025}, and instruction-tuned LLMs in clinical and biomedical NLP show similarly task dependent few-shot gains \cite{Labrak2024}. Comparative work on health-related text classification further supports evaluating traditional machine learning, supervised pretrained language models, and LLM prompting within a shared experimental framework \cite{Guo2024}. This motivates our comparison among TF-IDF baselines, BioBERT, and prompted LLMs on the same human-calibrated gold set, rather than treating LLM performance as self-evidently superior.

\subsection{LLMs for Triage and Urgency Evaluation}\label{llms-for-triage-and-urgency-evaluation}

LLM-based triage has recently emerged as an active research direction, particularly in emergency acuity assessment and patient-message urgency. \citet{Williams2024} evaluate LLMs for clinical acuity assessment in emergency department settings, \citet{Masanneck2024} compare LLM triage performance in emergency medicine, and large-scale retrospective ED evaluations further emphasize caution before autonomous deployment \cite{Nedos2026}. In the patient-message setting, \citet{Gatto2026} frames urgency as a pairwise ranking problem over asynchronous portal messages. Together, these studies establish that LLMs may be useful for triage-related tasks, but they also highlight that triage errors are clinically asymmetric, and more importantly under-triage of urgent or emergency cases can carry greater risk than conservative over-triage. Therefore, model evaluation should include safety-aware outcomes such as high-acuity recall, false-negative counts, under-triage, and severe under-triage, not only accuracy or macro-$F_1$.

\subsection{Positioning of This Paper}\label{positioning-of-this-paper}

Prior work has treated triage primarily as emergency acuity prediction \cite{Williams2024, Masanneck2024, Nedos2026} or as pairwise urgency ranking of patient messages \cite{Gatto2026}. We instead frame it as a four-class actionable routing task that asks what response a patient inquiry warrants, not what condition it indicates. We compare supervised TF-IDF and BioBERT baselines against prompted open-source and API LLMs on a single human-calibrated gold set, and we add three further analyses: prompt sensitivity, safety-aware metrics, and two-model consensus as a selective prediction signal. The resulting system is positioned as decision support for triage prioritization and selective human review, not as a replacement for clinician judgment.

\section{Data and Annotation}

\subsection{Source Dataset}\label{source-dataset}

We use the HealthCareMagic-100K dataset released with ChatDoctor \cite{Li2023}, a publicly available collection of 112,165 anonymized patient-physician consultation exchanges from HealthCareMagic.com. Each record contains a patient inquiry and a physician response. In this study, only the patient inquiry text is used as model input during evaluation. We select HealthCareMagic-100K because it contains broad consumer-health inquiries rather than being restricted to a single disease domain. This makes it more appropriate for general online triage categorization than disease-specific dialogue dataset such as CovidDialog, which focuses on COVID-19-related consultations \cite{Zhou2021}.

\subsection{Filtering and Preprocessing}\label{filtering-and-preprocessing}

We first apply a quality filter to the patient inquiry text. A record is excluded if any of the following conditions are met:

\begin{itemize}
    \item
      the patient message contains fewer than 20 tokens;
    \item
      the patient message contains more than 500 tokens;
    \item
      the raw patient-message length is below 10 characters.
\end{itemize}

This filtering step removes 2,002 records, leaving 110,163 usable patient messages. 

\subsection{Keyword-Stratified Sampling and Data Splits}\label{keyword-stratified-sampling-and-data-splits}

We expect high-acuity such as emergency-referral cases to be rare in online consultation data, so we construct a keyword-stratified working pool to increase the representation of potentially higher-acuity inquiries. Each record receives an emergency-enrichment score that rewards strong and moderate emergency keywords in the patient message and explicit escalation phrases in the physician response, while penalizing past-tense indicators that suggest retrospective rather than active concerns. Records are then assigned to keyword-derived buckets: self-care-enriched, schedule-visit-enriched, urgent-clinician-enriched and emergency-enriched, using priority ordering so that higher-confidence emergency signals take precedence. Within each bucket, records are sorted by priority score (random seed 42 for stochastic operations). Physician responses are used solely for enrichment and are excluded from all model inputs. The sampling target balanced coverage across buckets, not balanced final triage labels. Full bucket priority rules, score thresholds, and keyword lists are given in Appendix \ref{apn:sampling}.

\begin{table}[h]
\centering
\small
\renewcommand{\arraystretch}{1.5} % Provides professional vertical padding
\begin{tabularx}{\linewidth}{l X p{1.0 cm}} % l = Split, X = Role (flexible), p = N (fixed small)
\toprule
\textbf{Split} & \textbf{Role} & \textbf{N} \\ 
\midrule
Silver training set & Supervised model training using Claude Sonnet 4.5 silver labels & 700 \\ 
\midrule
Gold evaluation set & Held-out evaluation set with human-calibrated labels & 300 \\ 
\midrule
Few-shot set & High-confidence, human-calibrated examples used as in-context demonstrations & 40 \\ 
\bottomrule
\end{tabularx}
\caption{Data splits and their roles in training and evaluation.}
\label{tab:data-stats}
\end{table}
\begin{table*}[t] % the * makes it span two columns
\centering
\small
\renewcommand{\arraystretch}{1.2}
\begin{tabularx}{\linewidth}{l l p{4.3cm} p{4.3cm}}
\toprule
\textbf{Label} & \textbf{Action} & \textbf{Definition} & \textbf{Example Cues} \\ 
\midrule
self-care & Manage at home & Symptoms or questions that can reasonably be managed without clinical contact, including informational, retrospective, or low-risk concerns. & Mild, routine, self-limited, or already evaluated issues \\
\midrule
schedule-visit & Routine appointment & Requires non-urgent clinician assessment, follow-up, medication management, or referral. & Persistent symptoms, medication renewal, referral request, non-urgent abnormal results \\ 
\midrule
urgent-clinician-review & Review within 24–48h & Potentially serious, worsening, or concerning symptoms requiring timely clinical review, but not clearly requiring emergency care. & Worsening pain, infection signs, concerning new symptoms, non-sudden neurological symptoms \\ 
\midrule
emergency-referral & Emergency care & Potentially life-threatening or time-sensitive conditions requiring immediate emergency evaluation. & Severe chest pain, stroke-like symptoms, severe breathing difficulty, collapse, active suicidal plan \\
\bottomrule
\end{tabularx}
\caption{Triage label schema with definitions and example cues.}
\label{tab:triage_label}
\end{table*}

From this candidate pool, we select a 1,040-record working pool and split it into three disjoint subsets: a 700-record silver training set, a 300-record gold evaluation set, and a 40 record few-shot pool (Table \ref{tab:data-stats}). Because keyword derived buckets do not perfectly map to actionable triage labels, the final label distributions remain imbalanced after labeling and human calibration.

\subsection{Triage Label Schema}\label{triage-label-schema}

We formulate online patient inquiry triage as a four-class actionable classification task. Each inquiry receives exactly one of four labels: self-care, schedule-visit, urgent-clinician-review, or emergency-referral (Table \ref{tab:triage_label}). The goal is to assign an appropriate level of clinical response, not to diagnose the condition or recommend treatment.

The annotation guideline also allows an auxiliary insufficient\_info flag to capture missing or ambiguous clinical context. This flag is not treated as a fifth label. Final triage labels are restricted to the four actionable categories above. The full annotation guideline is provided in Appendix \ref{apn:additional-labeling-guideline-details}.

\subsection{Silver, Gold and Few-shot Label Construction}\label{silver-and-gold-label-construction}

We develop a structured annotation guideline for the four triage categories starting from a two-person pilot study on 50 held-out examples; both researchers label the pilot independently, then compare decisions and identify recurring disagreement patterns. The guideline is refined over six rounds: revisions add a ``triage from text alone'' principle, an informational versus active-symptom distinction, an insufficient-information flag, refined self-care criteria, tighter edge-case rules for special populations and laboratory-result questions, and explicit rule priority ordering. The final guideline includes label definitions, decision flowcharts, edge-case rules, and worked examples, and is used for the initial Claude Sonnet 4.5 labeling of the silver, gold, and few-shot subsets. Following this initial annotation, humans calibrate the gold and few-shot data.

All 1,040 working-pool records receive initial triage labels from Claude Sonnet 4.5 (temperature=1.0, output\_tokens=32,000), which produce structured JSON with predicted label, confidence and reasoning. The final label field is restricted to the four actionable triage categories. For the 700-record silver training set, these labels are used directly as silver supervision; the gold evaluation set and few-shot pool undergo additional human calibration before use.

\subsubsection{Gold, Few-shot Set Calibration}

The gold evaluation set consists of 300 records human-calibrated by two researchers. Both researchers independently review all examples using the finalized annotation guidelines; cases involving disagreement or uncertainty are discussed and adjudicated to produce the final gold labels.

\begin{table}[H]
\centering
% \footnotesize
\renewcommand{\arraystretch}{1.3}
\resizebox{\columnwidth}{!}{%
\begin{tabular}{l|c|c}
% \toprule
\hline
\textbf{Metric} & \textbf{Gold Eval (N=300)} & \textbf{Few-shot (N=40)} \\ 
\midrule
Labels retained & 186 (62.0\%) & 34 (85.0\%) \\ 
Labels revised & 114 (38.0\%) & 6 (15.0\%) \\ 
Sonnet–human Cohen’s $\kappa$ & 0.35 & 0.86 \\ 
% \bottomrule
\hline
\end{tabular}%
}
\caption{Human calibration outcomes for the gold and few-shot sets, relative to initial Claude Sonnet 4.5 silver labels.}
\label{tab:gold_cal}
\end{table}

\begin{table*}[t] % Adding the * makes it span the whole page width
\centering
\small
\renewcommand{\arraystretch}{1.4}
\begin{tabular}{l cc cc cc}
\toprule
& \multicolumn{2}{c}{\textbf{Silver Training}} & \multicolumn{2}{c}{\textbf{Gold Eval}} & \multicolumn{2}{c}{\textbf{Few-shot}} \\
\cmidrule(lr){2-3} \cmidrule(lr){4-5} \cmidrule(lr){6-7}
\textbf{Label} & \textbf{n} & \textbf{\%} & \textbf{n} & \textbf{\%} & \textbf{n} & \textbf{\%} \\ 
\midrule
self-care & 203 & 29.0 & 144 & 48.0 & 15 & 37.5 \\ 
schedule-visit & 214 & 30.6 & 93 & 31.0 & 11 & 27.5 \\ 
urgent-clinician-review & 192 & 27.4 & 37 & 12.3 & 9 & 22.5 \\ 
emergency-referral & 91 & 13.0 & 26 & 8.7 & 5 & 12.5 \\ 
\midrule
\textbf{Total} & \textbf{700} & \textbf{100.0} & \textbf{300} & \textbf{100.0} & \textbf{40} & \textbf{100.0} \\ 
\bottomrule
\end{tabular}
\caption{Final triage-label distribution across silver, gold, and few-shot splits.}
\label{tab:dataset_dis_after_cal}
\end{table*}

Compared with the initial Claude Sonnet 4.5 labels, 186 of 300 gold-set labels are retained (62.0\%), while 114 are revised (38.0\%). Agreement between the initial Sonnet labels and the final human-calibrated labels is \(\kappa = 0.35\), indicating fair agreement (Table~\ref{tab:gold_cal}). This value reflects Sonnet--human agreement rather than human--human inter-annotator agreement, and the relatively high revision rate highlights the need for human calibration when constructing the evaluation set.

A separate 40-record few-shot pool is selected from the 1,040-record working pool for prompt construction. These examples are chosen from high-confidence Claude Sonnet 4.5 labels and then human-verified as clear cases. Compared with the initial Sonnet labels, 34 of 40 labels are retained (85.0\%) and 6 are revised (15.0\%), with Sonnet--human agreement of \(\kappa = 0.86\) (Table~\ref{tab:gold_cal}). The higher agreement for the few-shot pool is expected because these examples are intentionally selected as high-confidence, representative demonstrations.

\subsection{Dataset Statistics}\label{dataset-statistics}
Table~\ref{tab:dataset_dis_after_cal} reports the final label distributions for the silver training set, gold evaluation set, and few-shot pool. The gold and few-shot distributions are based on labels after human calibration. Despite keyword-stratified sampling, all three datasets remain class-imbalanced, with emergency-referral cases less frequent than lower-acuity categories.

This imbalance motivates using macro-\(F_1\) as the primary metric, supplemented by safety-aware metrics including urgent recall, emergency recall, under-triage rate, and severe under-triage rate.

\section{Models and Experimental Setup}

\subsection{Experimental Design}\label{experimental-design}
We evaluate two modeling paradigms for four-class actionable triage classification: supervised classifiers and prompted LLM classifiers. Supervised models are trained on a 700-example silver-labeled training set generated by Claude Sonnet 4.5 and evaluated on the 300-example human-calibrated gold set. Prompted LLMs use fixed triage instructions to directly assign structured labels to the same gold-set inquiries.

Macro-$F_1$ is the primary evaluation metric. Accuracy, per-class $F_1$, and safety-aware triage metrics are reported as secondary analyzes. Table \ref{tab:model_family} summarizes the model families evaluated in this study. Full prompted-LLM model identifiers and sources are provided in Appendix \ref{apn:llm-source}.

\begin{table}[ht]
\centering
\scriptsize % Necessary for 2-column single-column fit
\setlength{\tabcolsep}{3pt} % Reduces horizontal padding
\renewcommand{\arraystretch}{1.6} % Keeps rows from looking squashed
\begin{tabular}{l p{2.8cm} l p{1.2cm}} % Manually tuned widths
\toprule
\textbf{Model group} & \textbf{Models} & \textbf{Mode} & \textbf{Training Dataset} \\ 
\midrule
TF-IDF & Logistic Regression, Random Forest, XGBoost & \makecell[l]{Train silver; \\ Eval gold} & Silver \\ 
\midrule
Bio-Encoder & BioBERT-v1.1 & \makecell[l]{Train silver; \\ Eval gold} & Silver \\ 
\midrule
Open LLMs & Llama3.1-8B, Qwen3-8B, Mistral-7B, DeepSeek-R1-7B & Prompted & None \\ 
\midrule
API LLMs & GPT-4o-mini, Haiku 4.5 & Prompted & None \\ 
\bottomrule
\end{tabular}
\caption{Model families evaluated. Supervised models train on silver labels; LLMs prompted without parameter updates.}
\label{tab:model_family}
\end{table}

\subsection{Classic Supervised Baselines}\label{Classic-supervised-baselines}

For each supervised model, we evaluate two training conditions. The default condition uses all 700 Claude Sonnet 4.5-generated silver-labeled examples. The balanced condition deterministically downsamples the silver set to 91 examples per class, yielding 364 total training examples.\footnote{The downsample size is determined by the emergency-referral class, which contains 91 examples in the silver-labeled training set; see Table~\ref{tab:dataset_dis_after_cal}.} The balanced variant is included to assess sensitivity to class imbalance, particularly for minority and safety-sensitive classes. Because downsampling also changes training-set size and composition, comparisons between default and balanced variants should be interpreted as sensitivity analyses rather than isolated causal estimates of class-imbalance effects. Additional supervised model details are provided in Appendix~\ref{apn:additional-supervised-details}.

\subsubsection{TF-IDF Classifiers}\label{tf-idf-classifiers}
The TF-IDF baseline is built using up to 5,000 unigram and bigram features. For each training condition, the vectorizer is fit only on the corresponding silver training subset and then applied to the held-out gold set. Using this representation, we train Logistic Regression (LR), Random Forest (RF), and eXtreme Gradient Boosting (XGBoost) classifiers. LR uses default regularization with max\_iter=1000; RF and XGBoost are tuned with 5-fold stratified cross-validation on the silver training subset, using macro-\(F_1\) as the selection criterion. After tuning, each model is refit on the same silver subset and used once to predict labels for the 300 gold-set. Full hyperparameter grids and selected values are provided in Appendix~\ref{apn:additional-supervised-details}.

\subsubsection{BioBERT Fine-Tuning}\label{biobert-fine-tuning}
The BioBERT supervised model is built by fine-tuning BioBERT-v1.1 (dmis-lab/biobert-v1.1) as a four-class sequence classifier~\cite{Lee2020}. Input texts are truncated or padded to 256 tokens, and a linear classification head is trained with cross-entropy loss. Fine-tuning uses 10 epochs, learning rate \(2 \times 10^{-5}\), batch size 16, weight decay 0.01, seed 42, and mixed-precision training on an NVIDIA T4 GPU. We evaluate both default and balanced BioBERT variants using the same supervised protocol as the TF-IDF models.

\subsection{Prompted LLM Classifiers}\label{prompted-llm-classifiers}

We evaluate six LLMs as in-context classifiers without parameter updates: Llama3.1-8B, Qwen3-8B, Mistral-7B, DeepSeek-R1-7B, GPT-4o-mini, and Claude Haiku 4.5. Full model configurations and inference details are provided in Appendix~\ref{apn:llm-source}.

% The four open-source models were served through Ollama on a Google Colab T4 runtime; GPT-4o-mini and Claude Haiku 4.5 were accessed through APIs. 
All LLMs use the same base triage prompt and are evaluated under three prompting settings: 0-shot, 4-shot, and 12-shot. The 0-shot setting uses only the base prompt; the 4-shot setting adds one example per class; and the 12-shot setting adds three examples per class, including the four examples used in the 4-shot setting. Full prompt and few-shot examples are provided in Appendix~\ref{apn:llm-shared-prompt-template}.

Few-shot examples are selected from the 40 record human-calibrated few-shot set reported in Table~\ref{tab:dataset_dis_after_cal}. They are selected as clear, high-confidence cases and used only for in-context learning, not for model training or final evaluation. 

\subsection{Output Parsing and Effective Evaluation Set}
\begin{table*}[ht]
\centering
\small
\setlength{\tabcolsep}{3.2pt}
\renewcommand{\arraystretch}{1.1}

\begin{tabular}{@{}llcccccc@{}}
\toprule
\textbf{Model family} 
& \textbf{Model / setting} 
& \textbf{Macro-\(F_1\) [95\% CI]} 
& \multicolumn{4}{c}{\textbf{Per-class \(F_1\)}} 
& \textbf{Parse-fail \%} \\
\cmidrule(lr){4-7}
& & 
& \makecell{\textbf{Self-care}} 
& \makecell{\textbf{Schedule}\\\textbf{visit}} 
& \makecell{\textbf{Urgent clinician}\\\textbf{review}} 
& \makecell{\textbf{Emergency}\\\textbf{referral}}
& \\
\midrule

TF-IDF & LR, default 
& 0.348 [0.292, 0.404] & 0.571 & 0.443 & 0.305 & 0.071 & 0.0 \\
TF-IDF & LR, balanced 
& 0.321 [0.267, 0.367] & 0.444 & 0.310 & 0.224 & 0.304 & 0.0 \\

\midrule
\addlinespace[2pt]

Transformer & BioBERT-v1.1, default 
& 0.378 [0.316, 0.435] & 0.573 & 0.425 & 0.252 & 0.264 & 0.0 \\
Transformer & BioBERT-v1.1, balanced 
& 0.307 [0.260, 0.356] & 0.533 & 0.294 & 0.168 & 0.233 & 0.0 \\

\midrule
\addlinespace[2pt]

LLM & GPT-4o-mini, 12-shot 
& 0.404 [0.351, 0.457] & 0.546 & 0.311 & 0.307 & 0.451 & 0.0 \\
LLM & Qwen3-8B, 4-shot 
& 0.444 [0.388, 0.498] & 0.655 & 0.408 & 0.296 & 0.417 & 0.7 \\
LLM & Llama3.1-8B, 12-shot 
& 0.464 [0.402, 0.520] & 0.664 & 0.477 & 0.336 & 0.379 & 1.0 \\
LLM & Claude Haiku 4.5, 12-shot 
& 0.475 [0.413, 0.532] & 0.532 & 0.469 & 0.341 & 0.557 & 0.0 \\

\bottomrule
\end{tabular}

\caption{Main classification results on the gold set. For each LLM, the best prompt setting for that model is shown. Macro-\(F_1\) is reported with 95\% bootstrap confidence intervals.}
\label{tab:model_performance}

\end{table*}
After label normalization, outputs that could not be mapped to one of the four valid triage labels are counted as parse failures and excluded from \(F_1\) and safety-metric computation. In Table~\ref{tab:model_performance}, Parse-fail \% reports the percentage of the 300 evaluation examples with invalid or unparseable labels. Supervised models produce one valid prediction for every example and therefore have a parse-fail rate of 0\%. Across LLM settings, parse failures are rare, with a maximum observed rate of 2.7\%.

For pairwise agreement analyses, metrics are computed on examples with valid predictions from both compared models. In consensus-based human-in-the-loop (HITL) simulations, invalid outputs from either model are treated as escalation to human review rather than auto-accepted predictions.

\subsection{Evaluation Metrics and Statistical Uncertainty}\label{evaluation-metrics-and-statistical-uncertainty}

We use macro-\(F_1\) as the primary metric to account for class imbalance and the clinical importance of high-acuity minority classes. Secondary metrics include per-class $F_1$ and safety-aware metrics. To quantify triage safety, we map the labels to an ordinal severity scale from 0 to 3: self-care, schedule-visit, urgent-clinician-review, and emergency-referral. Under-triage is defined as predicting a lower severity than the true label; severe under-triage is defined as a gap of at least two levels; and over-triage is defined as predicting a higher severity. We also report urgent and emergency recall.

We estimate 95\% confidence intervals for macro-$F_1$ using case-level bootstrap resampling with 1,000 replicates. Given the modest gold-set size and the number of model-prompt combinations, model comparisons are interpreted descriptively rather than as confirmatory hypothesis tests.

\section{Results}

We evaluate three model families: TF-IDF baselines, BioBERT-v1.1, and prompted LLMs on the 300-example gold set, using macro-$F_1$ as the primary metric and safety-aware metrics defined in section \ref{evaluation-metrics-and-statistical-uncertainty}.

\subsection{Main Classification Performance}

Table \ref{tab:model_performance} summarizes the main classification results. For supervised baselines, we report both default and balanced class-weight settings; for LLMs, we report the best prompt setting per model by macro-$F_1$.

Two findings emerge. First, supervised baselines perform unevenly across classes and are especially weak on emergency-referral: default-weight TF-IDF and BioBERT achieve moderate self-care performance but low emergency $F_1$, indicating that high-acuity minority classes are difficult under the default training distribution. Class re-weighting changes the per-class trade-off but does not improve overall macro-$F_1$. Balanced models gain on emergency in some cases but lose on more frequent classes, such as schedule-visit. The strongest supervised baseline is BioBERT-v1.1 with default weights at macro-$F_1$ of 0.378.

Second, the leading few-shot LLMs outperform all supervised baselines by point estimate: Claude Haiku 4.5 (12-shot) achieves the highest macro-$F_1$ at 0.475, followed by Llama3.1-8B (12-shot) at 0.464 and Qwen3-8B (4-shot) at 0.444. Confidence intervals overlap with the best supervised baseline, so this advantage should be interpreted as a favorable point-estimate trend rather than definitive statistical separation. Pairwise McNemar tests (Appendix \ref{apn:mcnemar}) reinforce this: the top three LLMs are not significantly different from one another, and only Llama3.1-8B reaches statistical significance against BioBERT-v1.1.

\subsection{Class-Specific and Safety-Aware Performance}

Aggregate macro-$F_1$ obscures two clinically important questions: which triage categories are reliably classified, and which models avoid clinically asymmetric errors.

Per-class performance is heterogeneous as shown in Table~\ref{tab:model_performance}. Self-care is the easiest class for leading LLMs: Llama3.1-8B (12-shot) and Qwen3-8B (4-shot) reach $F_1$ scores of 0.664 and 0.655. Schedule-visit shows moderate performance across the strongest configurations, ranging between 0.311 and 0.477. Urgent-clinician-review is the most consistently difficult class: no leading LLM exceeds urgent $F_1$ of 0.341, suggesting an intrinsically ambiguous intermediate-acuity boundary that is hard to infer from short patient text when clinical context is incomplete. Emergency-referral is more discriminative across models: Claude Haiku 4.5 (12-shot) achieves the highest $F_1$ at 0.557, followed by GPT-4o-mini (12-shot) at 0.451, Qwen3-8B (4-shot) at 0.417, and Llama3.1-8B (12-shot) at 0.379. These differences are clinically important because emergency-referral represents the highest risk category in the schema.
% \begin{table*}[ht]
% \centering
% \small
% \setlength{\tabcolsep}{3pt}
% \caption{Safety-aware metrics on the gold set with 95\% bootstrap CI. Under-triage is defined as predicting a lower-acuity label than the gold label. Severe under-triage is defined as an acuity gap of at least two levels.}
% \label{tab:safety_metrics}
% \begin{tabular}{llcccc}
% \hline
% \textbf{Model} & \textbf{Setting} & \textbf{Emergency recall} & \textbf{Urgent-or-higher recall} & \textbf{Under-triage rate} & \textbf{Severe under-triage rate} \\ \hline
% BioBERT-v1.1 & default & 0.269 [0.115, 0.462] & 0.698 [0.571, 0.810] & 0.137 [0.107, 0.167] & 0.308 [0.153, 0.500] \\
% TF-IDF + LR & balanced & 0.654 [0.462, 0.808] & 0.746 [0.635, 0.842] & 0.110 [0.077, 0.143] & 0.269 [0.115, 0.462] \\
% GPT-4o-mini & 12-shot & 0.615 [0.423, 0.808] & 0.984 [0.952, 1.000] & 0.053 [0.033, 0.077] & 0.000 [0.000, 0.000] \\
% Qwen3-8B & 4-shot & 0.577 [0.385, 0.769] & 0.937 [0.873, 0.984] & 0.090 [0.063, 0.120] & 0.000 [0.000, 0.000] \\
% Llama3.1-8B & 12-shot & 0.423 [0.231, 0.615] & 0.873 [0.794, 0.952] & 0.107 [0.077, 0.133] & 0.000 [0.000, 0.000] \\
% Claude Haiku 4.5 & 12-shot & 0.654 [0.462, 0.846] & 0.873 [0.794, 0.952] & 0.073 [0.047, 0.103] & 0.000 [0.000, 0.000] \\ \hline
% \end{tabular}
% \end{table*}
\begin{table*}[ht]
\centering
\small
\setlength{\tabcolsep}{4pt}
\renewcommand{\arraystretch}{1.12}

\begin{tabular}{@{}llcccc@{}}
\toprule
\textbf{Model} 
& \textbf{Setting} 
& \makecell{\textbf{Emergency}\\\textbf{recall}} 
& \makecell{\textbf{Urgent-or-higher}\\\textbf{recall}} 
& \makecell{\textbf{Under-triage}\\\textbf{rate}} 
& \makecell{\textbf{Severe under-triage}\\\textbf{rate}} \\
\midrule

BioBERT-v1.1 & default 
& 0.269 [0.115, 0.462] 
& 0.698 [0.571, 0.810] 
& 0.137 [0.107, 0.167] 
& 0.308 [0.153, 0.500] \\

TF-IDF + LR & balanced 
& 0.654 [0.462, 0.808] 
& 0.746 [0.635, 0.842] 
& 0.110 [0.077, 0.143] 
& 0.269 [0.115, 0.462] \\

\midrule
GPT-4o-mini & 12-shot 
& 0.615 [0.423, 0.808] 
& 0.984 [0.952, 1.000] 
& 0.053 [0.033, 0.077] 
& 0.000 [0.000, 0.000] \\

Qwen3-8B & 4-shot 
& 0.577 [0.385, 0.769] 
& 0.937 [0.873, 0.984] 
& 0.090 [0.063, 0.120] 
& 0.000 [0.000, 0.000] \\

Llama3.1-8B & 12-shot 
& 0.423 [0.231, 0.615] 
& 0.873 [0.794, 0.952] 
& 0.107 [0.077, 0.133] 
& 0.000 [0.000, 0.000] \\

Claude Haiku 4.5 & 12-shot 
& 0.654 [0.462, 0.846] 
& 0.873 [0.794, 0.952] 
& 0.073 [0.047, 0.103] 
& 0.000 [0.000, 0.000] \\

\bottomrule
\end{tabular}

\vspace{2pt}
\begin{minipage}{0.96\textwidth}
\footnotesize
\textit{Note:} Under-triage is defined as predicting a lower-acuity label than the gold label. Severe under-triage is defined as an acuity gap of at least two levels.
\end{minipage}
\caption{Safety-aware metrics on the gold set with 95\% bootstrap confidence intervals.}
\label{tab:safety_metrics}

\end{table*}

Safety-aware metrics (Table \ref{tab:safety_metrics}) reveal trade-offs that macro-$F_1$ alone hides. GPT-4o-mini (12-shot) does not match the top-performing LLMs in macro-$F_1$ but achieves the highest urgent-or-higher recall, 0.984 [0.952, 1.000], and the lowest under-triage rate, 0.053 [0.033, 0.077], among all evaluated configurations. This suggests that a model with lower aggregate $F_1$ may still be preferable in a workflow that prioritizes minimizing missed escalations. All four leading LLM configurations have an observed severe under-triage rate of 0.000 on the gold set, whereas supervised baselines show severe under-triage rates ranging from 0.269 to 0.308 under both default and balanced settings. This difference is clinically important but should be interpreted cautiously because the gold test set contains only 26\footnote{The gold test set includes 26 emergency-referral examples; see Table \ref{tab:dataset_dis_after_cal}.} emergency-referral examples. An observed zero rate in this sample is not a deployment-level safety guarantee. Rather, it indicates that safety-aware evaluation can reveal clinically meaningful model differences that aggregate metrics miss.

\begin{figure*}[t]
    \centering
    \includegraphics[width=0.85\textwidth]{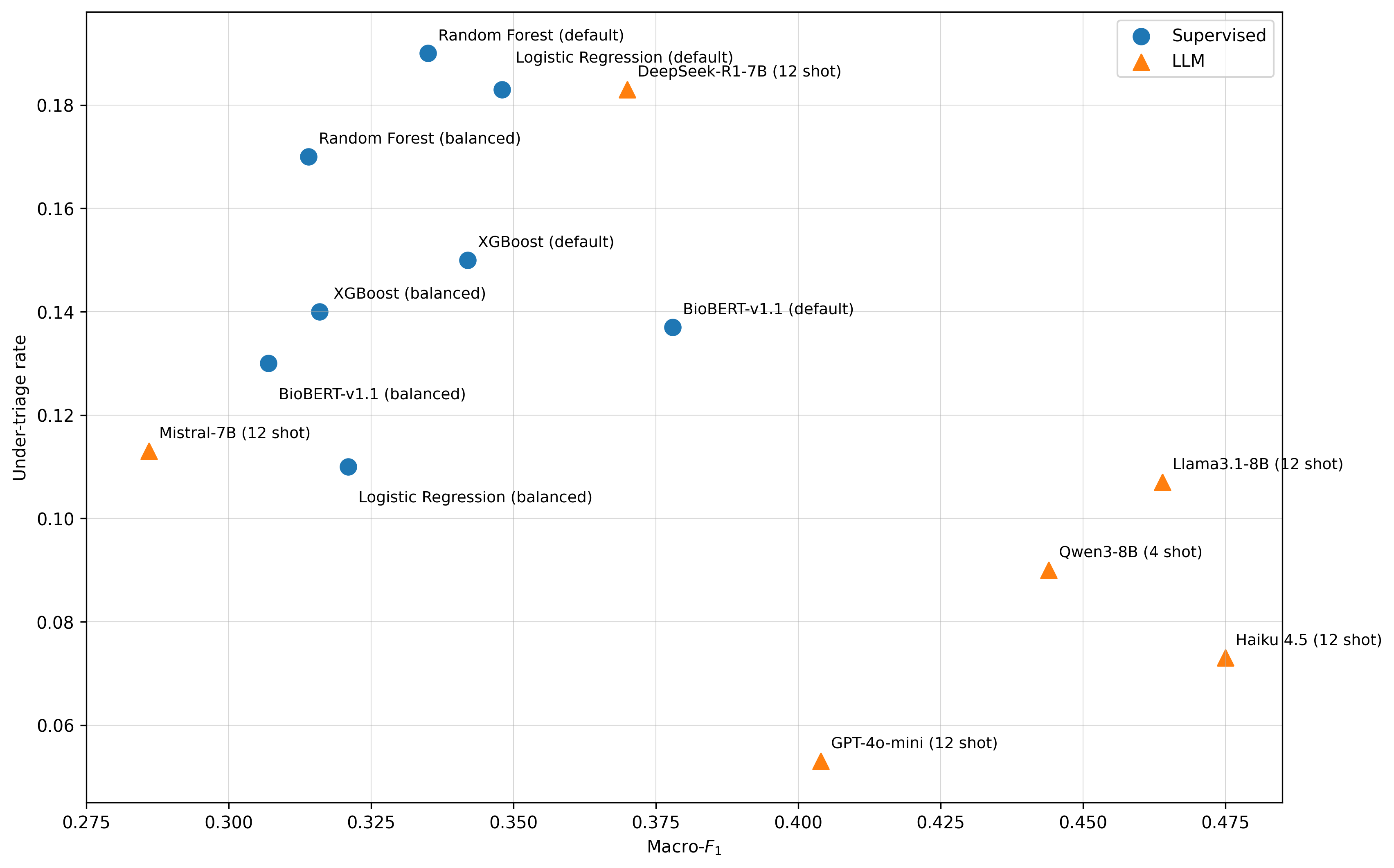}
    \caption{Trade-off between macro-$F_1$ (x-axis) and under-triage rate (y-axis) across model configurations. Few-shot LLMs occupy the upper-left region: higher macro-$F_1$ and lower under triage than supervised baselines.}
    \label{fig:perf_safety_tradeoff}
\end{figure*}

Figure \ref{fig:perf_safety_tradeoff} visualizes the trade-off between macro-$F_1$ and under-triage rate across configurations. The LLM configurations occupy a more favorable region of the plot, with higher macro-$F_1$ and generally lower under-triage than the supervised baselines.

\subsection{Prompt Sensitivity}
% \begin{table*}[ht]
% \centering
% \resizebox{\textwidth}{!}{%
% \begin{tabular}{lcccccc}
% \hline
% \textbf{Model} & \textbf{0-shot} & \textbf{4-shot} & \textbf{12-shot} & \textbf{\shortstack{Best\\setting}} & \textbf{\shortstack{$\Delta$ best\\vs 0-shot}} \\ \hline
% Claude Haiku 4.5 & 0.374 [0.318, 0.430] & 0.422 [0.365, 0.479] & \textbf{0.475 [0.413, 0.532]} & 12-shot & \textbf{+0.101} \\
% Llama3.1-8B & 0.375 [0.323, 0.426] & 0.351 [0.295, 0.405] & 0.464 [0.402, 0.520] & 12-shot & +0.089 \\
% Qwen3-8B & \textbf{0.385 [0.331, 0.439]} & \textbf{0.444 [0.388, 0.498]} & 0.437 [0.385, 0.488] & 4-shot & +0.058 \\
% GPT-4o-mini & 0.362 [0.310, 0.412] & 0.332 [0.281, 0.385] & 0.404 [0.351, 0.457] & 12-shot & +0.042 \\
% Mistral-7B & 0.260 [0.168, 0.295] & 0.246 [0.215, 0.279] & 0.286 [0.249, 0.323] & 12-shot & +0.027 \\
% DeepSeek-R1-7B & 0.350 [0.296, 0.402] & 0.342 [0.291, 0.391] & 0.370 [0.317, 0.420] & 12-shot & +0.020 \\ \hline
% Average (6 models) & 0.351 & 0.356 & 0.406 & — & +0.055 \\ \hline
% \end{tabular}%
% }
% \caption{Macro-$F_1$ across 0-,4-,12-shot prompting per LLM, with the best setting and improvement over 0-shot.}
% \label{tab:prompt_sensitivity}
% \end{table*}

\begin{table*}[ht]
\centering
\small
\setlength{\tabcolsep}{5pt}
\renewcommand{\arraystretch}{1.12}

\begin{tabular}{@{}lccccc@{}}
\toprule
\textbf{Model} 
& \textbf{0-shot} 
& \textbf{4-shot} 
& \textbf{12-shot} 
& \makecell{\textbf{Best}\\\textbf{setting}} 
& \makecell{\(\boldsymbol{\Delta}\) \textbf{best}\\\textbf{vs. 0-shot}} \\
\midrule

Claude Haiku 4.5 
& 0.374 [0.318, 0.430] 
& 0.422 [0.365, 0.479] 
& \textbf{0.475 [0.413, 0.532]} 
& 12-shot 
& \textbf{+0.101} \\

Llama3.1-8B 
& 0.375 [0.323, 0.426] 
& 0.351 [0.295, 0.405] 
& \textbf{0.464 [0.402, 0.520]} 
& 12-shot 
& +0.089 \\

Qwen3-8B 
& 0.385 [0.331, 0.439] 
& \textbf{0.444 [0.388, 0.498]} 
& 0.437 [0.385, 0.488] 
& 4-shot 
& +0.058 \\

GPT-4o-mini 
& 0.362 [0.310, 0.412] 
& 0.332 [0.281, 0.385] 
& \textbf{0.404 [0.351, 0.457]} 
& 12-shot 
& +0.042 \\

Mistral-7B 
& 0.260 [0.168, 0.295] 
& 0.246 [0.215, 0.279] 
& \textbf{0.286 [0.249, 0.323]} 
& 12-shot 
& +0.027 \\

DeepSeek-R1-7B 
& 0.350 [0.296, 0.402] 
& 0.342 [0.291, 0.391] 
& \textbf{0.370 [0.317, 0.420]} 
& 12-shot 
& +0.020 \\

\midrule
Average 
& 0.351 
& 0.356 
& 0.406 
& -- 
& +0.055 \\

\bottomrule
\end{tabular}

\vspace{2pt}
\begin{minipage}{0.96\textwidth}
\footnotesize
\textit{Note:} The average row reports the unweighted mean across the six LLMs.
\end{minipage}
\caption{Prompt sensitivity of LLMs on the gold set. Macro-\(F_1\) is reported with 95\% bootstrap confidence intervals. Bold values indicate the best prompt setting for each model.}
\label{tab:prompt_sensitivity}

\end{table*}
Prompt sensitivity is evaluated by comparing 0-shot, 4-shot, and 12-shot prompting for each LLM (Table \ref{tab:prompt_sensitivity}). Prompt demonstrations improve average performance -- macro-$F_1$ increases from 0.351 in the 0-shot setting to 0.356 in the 4-shot setting and 0.406 in the 12-shot setting -- but the effect is
neither monotonic nor uniform across models. Claude Haiku 4.5 improves monotonically and achieves its best performance at 12-shot. Llama3.1-8B also achieves its best performance at 12-shot, but its 4-shot score falls below 0-shot. Qwen3-8B achieves its best performance at 4-shot, indicating that additional demonstrations are not uniformly beneficial. GPT-4o-mini, Mistral-7B, and DeepSeek-R1-7B show non-monotonic or modest gains.

These findings indicate that the demonstration count alone is an insufficient design variable. Few-shot prompting should be treated as model-specific calibration rather than a uniformly beneficial intervention. In triage settings, prompt selection should also consider safety-aware behavior, because a prompt that improves macro-$F_1$ may still change the balance between under-triage and over-triage.

\subsection{Model Disagreement and HITL Analysis}

Single model results motivate a selective prediction analysis -- rather than committing every case to one model, a triage workflow could auto-accept predictions only when two models agree and route the rest to human review. We evaluate this strategy as an oracle-HITL simulation. For each model pair, the consensus rule is: if both models produce the same valid label, the shared label is auto-accepted; otherwise, the case is escalated to human review. We report the metrics below: 

\begin{itemize}
    \item 
        Best single-model macro-\(F_1\) among the two models; manual escalation rate - the fraction of cases where the two models disagree and send them for human review.
    \item
      Consensus accuracy and consensus-only macro-$F_1$ — performance computed only on the auto-accepted subset where both models agree.
    \item
      Oracle-HITL macro-$F_1$ — full-cohort macro-$F_1$ under the
      assumption that all escalated cases receive the correct gold label.
\end{itemize}

The oracle-HITL metric is an upper bound on a selective review workflow, not an estimate of real-world clinician performance. Consensus-only metrics are also not directly comparable to full-cohort metrics because the auto-accepted subset is selected by model agreement and is likely to over-represent easier cases.

\begin{table*}[ht]
\centering
\small
\setlength{\tabcolsep}{4pt}
\renewcommand{\arraystretch}{1.15}

\begin{tabular}{@{}clcccccc@{}}
\toprule
\textbf{ID}
& \textbf{Model pair} 
& \makecell{\textbf{Best single}\\\textbf{macro-\(F_1\)}} 
& \makecell{\textbf{Manual esc.}\\\textbf{rate}} 
& \makecell{\textbf{Consensus}\\\textbf{accuracy}} 
& \makecell{\textbf{Consensus-only}\\\textbf{macro-\(F_1\)}} 
& \makecell{\textbf{Oracle-HITL}\\\textbf{macro-\(F_1\)}} 
& \makecell{\(\boldsymbol{\Delta}\)\\\textbf{macro-\(F_1\)}} \\
\midrule

P1 
& \makecell[l]{GPT-4o-mini 12-shot +\\ Llama3.1-8B 12-shot}
& 0.464
& \makecell{42.7\%\\{[37.0, 48.3]}}
& \makecell{56.4\%\\{[49.1, 63.5]}}
& \makecell{0.538\\{[0.458, 0.612]}}
& \makecell{\textbf{0.708}\\\textbf{[0.650, 0.769]}}
& \makecell{\textbf{+0.244}\\\textbf{[0.190, 0.300]}} \\

\addlinespace[4pt]

P2 
& \makecell[l]{Claude Haiku 4.5 12-shot +\\ Llama3.1-8B 12-shot}
& \textbf{0.475}
& \makecell{39.0\%\\{[33.0, 44.3]}}
& \makecell{56.8\%\\{[49.2, 63.9]}}
& \makecell{\textbf{0.550}\\\textbf{[0.464, 0.618]}}
& \makecell{0.701\\{[0.642, 0.756]}}
& \makecell{+0.226\\{[0.177, 0.261]}} \\

\addlinespace[4pt]

P3 
& \makecell[l]{Llama3.1-8B 12-shot +\\ Qwen3-8B 4-shot}
& 0.464
& \makecell{36.0\%\\{[30.3, 41.0]}}
& \makecell{\textbf{59.9\%}\\\textbf{[53.0, 66.1]}}
& \makecell{0.537\\{[0.460, 0.609]}}
& \makecell{0.691\\{[0.635, 0.744]}}
& \makecell{+0.227\\{[0.174, 0.269]}} \\

\addlinespace[4pt]

P4 
& \makecell[l]{Claude Haiku 4.5 12-shot +\\ Qwen3-8B 4-shot}
& \textbf{0.475}
& \makecell{35.0\%\\{[30.0, 40.3]}}
& \makecell{53.8\%\\{[46.2, 60.4]}}
& \makecell{0.520\\{[0.443, 0.588]}}
& \makecell{0.655\\{[0.592, 0.713]}}
& \makecell{+0.179\\{[0.141, 0.213]}} \\

\addlinespace[4pt]

P5 
& \makecell[l]{Llama3.1-8B 12-shot +\\ Qwen3-8B 12-shot}
& 0.464
& \makecell{\textbf{32.0\%}\\\textbf{[26.7, 37.0]}}
& \makecell{58.8\%\\{[52.0, 65.1]}}
& \makecell{0.531\\{[0.453, 0.603]}}
& \makecell{0.663\\{[0.601, 0.720]}}
& \makecell{+0.199\\{[0.150, 0.243]}} \\

\bottomrule
\end{tabular}
\caption{Two-model consensus analysis on the gold set. Manual escalation rate corresponds to the disagreement rate between the two models. Oracle-HITL macro-\(F_1\) assumes that manually escalated disagreement cases are resolved correctly.}
\label{tab:model_pairs_evaluation}

\end{table*}

Across the five evaluated model pairs (Table~\ref{tab:model_pairs_evaluation}), manual escalation rates range from 32.0\% to 42.7\%. Under the oracle-HITL assumption, all five pairs substantially exceed the stronger single-model macro-\(F_1\) in each pair. The highest oracle-HITL macro-$F_1$ is achieved by GPT-4o-mini (12-shot) + Llama3.1-8B (12-shot) (0.708 {[}0.650, 0.769{]}); the highest consensus-only macro-$F_1$ is Claude Haiku 4.5 (12-shot) + Llama3.1-8B (12-shot) (0.550 {[}0.464, 0.618{]}). A practically notable pair is Llama3.1-8B (12-shot) + Qwen3-8B (4-shot), which achieves the highest consensus accuracy in our evaluation (59.9\% {[}53.0, 66.1{]}) with a moderate manual escalation rate (36.0\% {[}30.3, 41.0{]}), and a high oracle-HITL macro-$F_1$ upper bound (0.691 {[}0.635, 0.744{]}) -- a useful candidate for future clinician-in-the-loop validation where cost, privacy, or deployment constraints favor locally hosted inference.

Per-class oracle-HITL $F_1$ and label-conditional consensus accuracy (Appendix \ref{apn:HITL}) show that consensus reliability is strongly label-dependent: agreement on self-care is highly reliable, whereas agreement on urgent-clinician-review is unreliable across all five evaluated pairs. In practice, this supports a selective prediction policy rather than autonomous triage -- self-care agreement may be suitable for auto-acceptance under audit, while urgent and emergency labels should remain subject to human review. We do not advocate autonomous deployment of any configuration evaluated here; the role evaluated is decision support with human oversight on high-acuity decisions.
\section{Discussion}
These results suggest that few-shot LLMs can provide useful decision-support signals for online patient-inquiry triage, but their value depends less on aggregate macro-\(F_1\) alone than on class-specific reliability, safety-aware error patterns, and workflow integration. The central finding is not that prompted LLMs are ready for autonomous triage, but that they may support selective prioritization when paired with human review and label-aware safeguards.

% \subsection{Summary of Main Findings}\label{summary-of-main-findings}

% This study evaluates four-class actionable triage classification for online patient inquiries using a human-calibrated gold set, supervised baselines, prompted LLMs, and safety-aware metrics. Three findings stand out.

% First, few-shot LLMs showed promising performance relative to supervised baselines, but the absolute macro-$F_1$ values remained modest. Claude Haiku 4.5, Llama3.1-8B, and Qwen3-8B achieved the strongest macro-$F_1$ point estimates, while BioBERT-v1.1 was the strongest supervised baseline. Because confidence intervals overlapped across several leading configurations, these results should be interpreted as evidence of relative promise rather than deployment-ready reliability.

% Second, performance varied substantially by class. Self-care was generally easier for leading models, whereas urgent-clinician-review remained consistently difficult. This suggests that intermediate-acuity triage is a central challenge for text-only classification, especially when patient-authored messages omit duration, severity, risk factors, or current care status.

% Third, safety-aware metrics revealed trade-offs that macro-$F_1$ alone would miss. Models differed in how they balanced under-triage and over-triage, and the consensus analysis showed that model agreement is useful only in a label-dependent way. Agreement on self-care was much more reliable than agreement on urgent-clinician-review, arguing against a uniform ``agree and auto-accept'' rule.

\subsection{Annotation Difficulty and Model Difficulty}

A consistent pattern emerges across annotation and evaluation: labels that required more guideline refinement are also harder for models to classify reliably. During guideline development, many disagreements center on boundaries such as self-care versus schedule-visit and schedule-visit versus urgent-clinician-review. These boundaries require deciding whether a patient-authored message is informational, requires non-urgent follow-up, or signals a need for timely clinical review.

The model results show a similar pattern. Urgent-clinician-review has low per-class $F_1$ across leading LLM configurations and is also the least reliable label under two-model consensus: when two models agree on urgent-clinician-review, the agreed label is correct only about one quarter of the time among the evaluated pairs. This is clinically plausible because urgent category occupies an ambiguous region of the severity spectrum: it is more concerning than routine follow-up but not clearly emergent. Such judgments often depend on information that is underspecified in online inquiries (symptom trajectory, severity, patient risk factors, prior evaluation, and current care status). 

This interpretation is hypothesis-generating, because the current design cannot fully separate annotation ambiguity, missing clinical context, model limitations, prompt design, and label-taxonomy effects.

\subsection{Interpretation of LLM Performance}\label{interpretation-of-llm-performance}

Several factors may explain why prompted LLMs outperform supervised baselines by macro-$F_1$ point estimate. First, LLMs can use pretrained linguistic and medical knowledge to interpret free-text symptom descriptions without task-specific parameter updates, while the supervised baselines are trained with only 700 silver-labeled examples (364 examples in the balanced-silver setting). Second, the supervised baselines are trained on Claude Sonnet 4.5 generated silver labels rather than human-calibrated labels. These labels provide a practical low-resource training signal, but they also introduce noise. The gold calibration results show that a substantial fraction of initial Sonnet labels are revised during human review, reinforcing that uncalibrated silver labels should not be treated as ground truth. Third, prompted LLMs receive explicit label definitions and decision rules derived from the annotation guideline. Few-shot examples also provide clear in-context anchors for the four labels. However, these demonstrations are selected to be high-confidence and unambiguous, not to represent the full range of difficult boundary cases. This may help explain why urgent-clinician-review remains difficult despite few-shot prompting.

Prompt sensitivity further complicates interpretation: few-shot performance reflects model capability, prompt design, and example selection rather than model capability alone. In this setting, prompt optimization should not be guided only by macro-\(F_1\), but also by under-triage, high-acuity recall, and label-specific reliability.

\subsection{Safety and Workflow Implications}\label{clinical-and-workflow-implications}

The results do not support autonomous deployment of any evaluated model for clinical triage. Even the strongest configurations misclassify a substantial fraction of examples, and urgent-clinician-review remains unreliable. These systems should therefore be viewed as potential decision-support components, not replacements for clinical judgment.

A more appropriate workflow is selective human review: LLMs may help prioritize messages, identify likely low-risk inquiries, flag possible high-acuity concerns, and route uncertain or disagreement cases to human review. The two-model consensus analysis suggests that agreement can identify a higher reliability subset, but only in a label-dependent way --- self-care agreement is relatively reliable, urgent-clinician-review agreement is not, and emergency-referral agreement requires caution because the clinical cost of a missed emergency is high and the gold set contains only 26 emergency cases. Low-risk agreement patterns, such as self-care agreement, could be evaluated for auto-acceptance under audit in future systems; urgent and emergency labels should remain subject to human review regardless of model agreement; and disagreements, invalid outputs, and ambiguous cases should be escalated rather than silently resolved.

The consensus results should be interpreted as workflow simulation rather than deployment evidence. The oracle-HITL metric assumes that every escalated case receives the correct gold label, making it an upper bound on a selective-review workflow. Real-world clinician review may be less consistent, especially because escalated cases are enriched for ambiguity. Prospective validation is required before making claims about workload reduction or safety.

Overall, these findings extend prior triage and clinical NLP work by shifting the evaluation focus from aggregate classification performance to workflow-level reliability. In this setting, the central question is not only which model achieves the highest macro-\(F_1\), but whether its errors, confidence signals, and disagreement patterns are compatible with safe human-supervised triage prioritization. This framing is especially important for patient-authored inquiries, where incomplete clinical context and asymmetric safety risks make autonomous routing inappropriate.

% \subsection{Relation to Prior Work and Contribution}

% Our findings align with three threads in prior work: clinical text features and supervised baselines are useful for triage but limited under small or noisy labels \cite{Ivanov2021, Si2020, Porto2024}. LLM prompting is task-dependent and prompt-sensitive \cite{Labrak2024, Sivarajkumar2024, Chen2025}. And LLM triage requires safety-aware evaluation rather than aggreagte accuracy alone \cite{Williams2024, Masanneck2024, Nedos2026, Lee2025}. The main distinction of this work is its workflow-level framing. Rather than evaluating triage only as emergency acuity prediction, urgency ranking, or general medical text classification, we frame the task as four-class actionable routing of noisy patient-authored inquiries, compare supervised baselines, BioBERT, and prompted LLMs under the same human-calibrated gold set, then evaluate prompt sensitivity, safety-aware errors, and oracle-HITL consensus analyzes. This shifts the central question from which model has the highest macro-$F_1$ to whether model behavior is safe, interpretable, and useful for human-supervised triage prioritization.

\section{Limitations}

Several limitations should be considered when interpreting these results.

First, this study uses a single public medical dialogue corpus. HealthCareMagic-100K provides broad online patient inquiry data, but it may not reflect patient portal messages, telemedicine chat, emergency department triage notes, or non-English clinical communication. External validation on additional patient-authored datasets and clinical settings is needed before drawing broader conclusions about generalizability.

Second, the gold evaluation set is modest in size. Although the 300-example gold set is human-calibrated, the emergency-referral class contains only 26 examples. As a result, emergency recall, severe under-triage, and emergency false-negative estimates have substantial uncertainty. The absence of observed severe under-triage among selected leading LLM configurations should therefore be interpreted as an encouraging signal rather than a safety guarantee.

Third, the gold labels are guideline informed and human calibrated, but not clinician adjudicated. Two human reviewers independently review the gold set and resolve disagreements through adjudication, but individual pre-adjudication labels are not preserved in a form that supports human-human inter-annotator agreement analysis. We therefore report Sonnet--human agreement rather than human-human \(\kappa\). Future work should include preserved independent annotations and, ideally, clinician adjudication for a larger subset.

Fourth, supervised baselines are trained on silver labels generated by Claude Sonnet 4.5 rather than on fully human-labeled training data. This reflects a realistic low-resource setting, but it also introduces label noise into the supervised training set. As a result, the supervised baselines should be interpreted as silver-supervised baselines rather than as upper-bound fully supervised models.

Fifth, prompted LLM performance depends on prompt design and few-shot example selection. The few-shot examples are selected to be high-confidence, human-calibrated, and relatively unambiguous, which makes them useful as clear in-context demonstrations but may underrepresent difficult boundary cases. Different prompt templates, example ordering, or intentionally ambiguous demonstrations could produce different results.

Sixth, the HITL analysis is an oracle simulation. The oracle-HITL metric assumes that every escalated case receives the correct gold label, so it represents an upper bound on a selective-review workflow rather than expected real-world clinical performance. Real clinical review may be less consistent, especially because escalated cases are enriched for disagreement and ambiguity. Prospective validation with clinician reviewers is required before using these results to estimate workload reduction or safety in practice.

Finally, this study evaluates models offline rather than in a deployed clinical workflow. Real-world use would introduce additional challenges, including distribution shift, changing patient behavior, latency constraints, user-interface design, escalation protocols, and institutional risk tolerance. We therefore view the results as evidence for triage decision support and prioritization research, not as evidence for autonomous clinical deployment.
\section{Conclusion}

We evaluate a four-class actionable triage classification of online patient inquiries, comparing supervised TF-IDF and BioBERT baselines against six prompted LLMs under 0-shot, 4-shot, and 12-shot conditions on a 300-example human-calibrated gold set. Few-shot LLMs achieve the strongest macro-$F_1$ point estimates, with Claude Haiku 4.5 (12-shot) reaching 0.475, but absolute performance remains modest and confidence intervals overlap with the best supervised baseline. Performance varies by class and by prompt setting, and urgent-clinician-review is consistently the hardest category for every model family. Safety-aware metrics reveal trade-offs that macro-$F_1$ alone misses -- the model with the highest macro-$F_1$ does not always have the lowest under-triage rate, and two-model agreement is a reliable confidence signal for self-care but not for urgent or emergency cases. Together, these results support using LLMs as decision support, with human oversight retain for high-acuity cases, rather than as autonomous triage. Future work should include clinician-adjusted labels, large gold sets with preserved independent annotations, and prospective evaluation in clinical workflows.

\bibliography{Triage}

\newpage
\appendix
\section{Additional Data Details}

\subsection{Keyword-Stratified Sampling}
\label{apn:sampling}

We used keyword-stratified sampling to enrich the candidate pool for higher-acuity inquiries, which are less frequent in online medical consultation data. Before sampling, we excluded records whose patient message contained fewer than 20 tokens, more than 500 tokens, or fewer than 10 raw characters. This removed 2,002 records (1.8\%), leaving 110,163 usable patient messages.

\paragraph{Emergency-enrichment scoring.}
Each record received an emergency-enrichment score based on keyword matches in the patient message and escalation language in the doctor response. Strong emergency keywords, moderate emergency keywords, and doctor escalation phrases increased the score, while past-tense indicators reduced priority because they often reflected retrospective rather than active emergency concerns. Records with an emergency score \(\geq 2\) were assigned to the emergency-enriched bucket. Records with emergency score \(=1\) were retained as lower-priority emergency-enriched candidates when bucket capacity remained.

\paragraph{Bucket assignment and selection.}
Records were assigned to sampling buckets using ordered keyword rules, so that emergency signals took precedence over lower-acuity keyword matches. The buckets and maximum selected candidate counts were:

\begin{enumerate}
    \item \textbf{Emergency-enriched}: emergency score \(\geq 2\), up to 1,200 candidates.
    \item \textbf{Self-care}: self-care keywords present and no self-care exclusion keywords, up to 800 candidates.
    \item \textbf{Urgent}: urgent keywords present, up to 500 candidates.
    \item \textbf{Schedule}: schedule, chronic-care, follow-up, or appointment-related keywords present, up to 500 candidates.
\end{enumerate}

Within each bucket, records were sorted by priority score before selection. Random seed 42 was used for stochastic operations.

\paragraph{Keyword lists.}
The following keyword lists were used for emergency enrichment and bucket assignment.

\begingroup
\small
\setlength{\parindent}{0pt}
\setlength{\parskip}{0.6em}
\sloppy

\textbf{\texttt{EMERGENCY\_KEYWORDS\_STRONG}:}
``can't breathe'', ``cannot breathe'', ``can not breathe'', ``struggling to breathe'', ``gasping for air'', ``not breathing'', ``stopped breathing'', ``just collapsed'', ``just passed out'', ``just fainted'', ``unresponsive'', ``won't wake up'', ``not responding'', ``having a seizure'', ``just had a seizure'', ``convulsing'', ``kill myself'', ``suicide'', ``want to die'', ``end my life'', ``overdose'', ``overdosed'', ``took too many pills'', ``choking'', ``can't swallow'', ``bleeding won't stop'', ``bleeding heavily'', ``severe bleeding'', ``heart attack'', ``crushing chest pain''.

\textbf{\texttt{EMERGENCY\_KEYWORDS\_MODERATE}:}
``chest pain'', ``difficulty breathing'', ``shortness of breath'', ``unconscious'', ``collapsed'', ``seizure'', ``passed out'', ``heavy bleeding'', ``stroke'', ``anaphylaxis'', ``coma'', ``can't move'', ``paralyzed'', ``severe allergic''.

\textbf{\texttt{PAST\_TENSE\_INDICATORS}:}
``had'', ``was'', ``ago'', ``last year'', ``last month'', ``last week'', ``few months ago'', ``years ago'', ``used to'', ``history of'', ``previously'', ``in the past'', ``recovered'', ``went to er'', ``went to the hospital'', ``was diagnosed''.

\textbf{\texttt{DOCTOR\_EMERGENCY\_PHRASES}:}
``go to er'', ``go to the er'', ``emergency room'', ``call 911'', ``call emergency'', ``go to hospital immediately'', ``seek immediate'', ``life-threatening'', ``go to the nearest'', ``immediately go'', ``rush to'', ``don't wait'', ``call an ambulance'', ``needs immediate attention''.

\textbf{\texttt{SELFCARE\_KEYWORDS}:}
``is this normal'', ``is it normal'', ``should i worry'', ``home remedy'', ``home remedies'', ``over the counter'', ``otc'', ``mild'', ``minor'', ``slight cold'', ``common cold'', ``vitamin'', ``nutrition'', ``diet'', ``supplement'', ``how long does'', ``will it go away'', ``go away on its own'', ``is it safe to'', ``can i take'', ``what can i do at home''.

\textbf{\texttt{SELFCARE\_EXCLUDERS}:}
``severe'', ``worst'', ``unbearable'', ``excruciating'', ``emergency'', ``can't breathe'', ``chest pain'', ``bleeding'', ``unconscious'', ``getting worse'', ``worsening'', ``spreading''.

\textbf{\texttt{URGENT\_KEYWORDS}:}
``getting worse'', ``worsening'', ``severe pain'', ``intense pain'', ``high fever'', ``blood in'', ``infection'', ``infected'', ``swelling'', ``swollen'', ``pus'', ``abscess'', ``lump'', ``can't sleep'', ``unable to eat'', ``unable to walk'', ``spreading'', ``excruciating'', ``unbearable'', ``not healing'', ``keeps coming back''.

\textbf{\texttt{SCHEDULE\_KEYWORDS}:}
``for weeks'', ``for months'', ``persistent'', ``recurring'', ``follow up'', ``follow-up'', ``medication'', ``prescription'', ``chronic'', ``diagnosed with'', ``specialist'', ``referral'', ``second opinion'', ``test results'', ``lab results'', ``been having'', ``for a while'', ``on and off'', ``appointment'', ``check up'', ``check-up''.

\endgroup

\section{Additional Supervised Model Details}
\label{apn:additional-supervised-details}

\subsection{TF-IDF Classifier Setup}
\label{apn:tfidf-classifier-setup}

Classical supervised baselines used a TF-IDF representation with up to 5,000 unigram and bigram features. For each supervised training condition, the TF-IDF vectorizer was fit only on the corresponding silver training subset and then applied to the held-out gold set. The vectorizer was never fit on gold text.

We trained three TF-IDF classifiers: Logistic Regression, Random Forest, and XGBoost. Logistic Regression used default regularization with max\_iter=1000 and random\_state=42. Random Forest and XGBoost hyperparameters were selected using 5-fold stratified cross-validation on the corresponding silver training subset, with macro-\(F_1\) as the selection criterion. After hyperparameter selection, the selected estimator was refit on the full corresponding silver subset and used to predict the gold set once.

\subsection{Random Forest and XGBoost Hyperparameter Search}
\label{apn:rf-xgb-hyperparameter-search}

Random Forest and XGBoost were tuned separately for the full-silver and balanced-silver training conditions. Table~\ref{tab:hyperparams_classical} summarizes the search grids and selected configurations for both training conditions. After hyperparameter selection, the selected estimator was refit on the corresponding silver subset and used to predict the held-out gold set once.

\subsection{BioBERT Fine-Tuning}
\label{apn:biobert-finetuning}

\begin{table}[t]
\centering
\small
\renewcommand{\arraystretch}{1.15}
\begin{tabular}{@{}ll@{}}
\toprule
\textbf{Configuration} & \textbf{Value} \\
\midrule
Base model & dmis-lab/biobert-v1.1 \\
Optimizer & AdamW \\
Learning rate (LR) & \(2 \times 10^{-5}\) \\
LR schedule & Linear, no warmup (trainer default) \\
Batch size & 16 train, 32 eval \\
Max sequence length & 256 tokens \\
Training epochs & 10 \\
Weight decay & 0.01 \\
Mixed precision & FP16 on NVIDIA T4 GPU \\
Random seed & 42 \\
Evaluation during training & None \\
Checkpoint selection & Final epoch checkpoint \\
\bottomrule
\end{tabular}
\caption{BioBERT fine-tuning configuration.}
\label{tab:biobert_params}
\end{table}
\begin{table*}[t]
\centering
\small
\renewcommand{\arraystretch}{1.15}
\begin{tabular*}{\textwidth}{@{\extracolsep{\fill}}llll@{}}
\toprule
\textbf{Model / Parameter} & \textbf{Search Grid} & \textbf{Full-Silver} & \textbf{Balanced-Silver} \\
\midrule
Random Forest & & & \\
n\_estimators & \{20, 50, 100, 200\} & 200 & 50 \\
max\_depth & \{3, 5, 10, 20, None\} & 20 & None \\
min\_samples\_leaf & \{1, 2, 3, 5, 10\} & 1 & 3 \\
max\_features & \{sqrt, log2\} & sqrt & sqrt \\
\midrule
XGBoost & & & \\
n\_estimators & \{100, 200, 300\} & 300 & 100 \\
max\_depth & \{4, 6, 8\} & 4 & 4 \\
learning\_rate & \{0.05, 0.1\} & 0.1 & 0.05 \\
min\_child\_weight & \{1, 3\} & 1 & 3 \\
subsample & \{0.8, 1.0\} & 1.0 & 0.8 \\
colsample\_bytree & \{0.6, 0.8, 1.0\} & 0.8 & 0.8 \\
\bottomrule
\end{tabular*}
\caption{Hyperparameter search space and selected configurations for classical supervised baselines. Search grids are shown compactly; selected configurations were chosen by 5-fold stratified cross-validation on the corresponding silver training subset using macro-\(F_1\).}
\label{tab:hyperparams_classical}
\end{table*}

We fine-tuned dmis-lab/biobert-v1.1 as a four-class sequence classifier. The model was loaded with a randomly initialized linear classification head over the four triage labels, and the encoder and classification head were fine-tuned jointly using cross-entropy loss.

Both the full-silver and balanced-silver BioBERT variants followed the same fixed training protocol. The full-silver variant was trained on all 700 silver-labeled examples, and the balanced-silver variant was trained on the 364-example class-balanced silver subset. No gold examples were used during fine-tuning.

We used the same fixed protocol for both BioBERT variants, as detailed in Table~\ref{tab:biobert_params}. No validation-based early stopping or checkpoint selection was performed. After the fixed 10-epoch fine-tuning schedule, the final checkpoint was used to predict the held-out 300-example gold test set once. Gold labels were not used for training, validation, early stopping, checkpoint selection, hyperparameter selection, or calibration.

\FloatBarrier

\section{Additional Labeling Guideline Details}
\label{apn:additional-labeling-guideline-details}

\subsection{Task definition}
\label{apn:annotation-task-definition}

Given a patient's online medical inquiry, annotators assigned exactly one of four actionable triage categories: self-care, schedule-visit, urgent-clinician-review, or emergency-referral. The goal was to determine the appropriate level of clinical response, not to diagnose the condition or recommend treatment.

The guideline also allowed an auxiliary insufficient-info flag to indicate that important clinical context was missing. This flag was not treated as a fifth final class. All final triage labels were restricted to the four actionable categories listed above.

\subsection{Label definitions}
\label{apn:annotation-label-definitions}

Table~\ref{tab:appendix_label_definitions} summarizes the four actionable triage labels, their definitions, and expected response timeframes.

% \begin{table}[H]
% \centering
% \renewcommand{\arraystretch}{1.3}
% \resizebox{\columnwidth}{!}{%
% \begin{tabular}{>{\raggedright\arraybackslash}p{0.30\columnwidth}|>{\raggedright\arraybackslash}p{0.48\columnwidth}|>{\raggedright\arraybackslash}p{0.16\columnwidth}}
% \hline
% \textbf{Label} & \textbf{Definition} & \textbf{Expected response timeframe} \\
% \midrule
% self-care & Manageable at home without clinical input; includes informational, retrospective, or low-risk questions without active concerning symptoms. & No clinical response needed \\
% schedule-visit & Needs clinical assessment but is not urgent; includes persistent symptoms, medication/referral needs, or non-urgent follow-up questions. & Days to weeks \\
% urgent-clinician-review & Requires timely clinical response due to worsening, moderate/severe, infectious, neurological, or special-population concerns that are not clearly emergent. & Within 24--48 hours \\
% emergency-referral & Potentially life-threatening or time-critical presentation requiring immediate emergency care. & Immediate \\
% \hline
% \end{tabular}%
% }
% \caption{Definitions and expected response timeframes for the four actionable triage labels.}
% \label{tab:appendix_label_definitions}
% \end{table}
\begin{table}[H]
\centering
\small
\renewcommand{\arraystretch}{1.18}
\begin{tabularx}{\columnwidth}{@{}>{\raggedright\arraybackslash}p{0.15\columnwidth}
                                >{\raggedright\arraybackslash}p{0.2\columnwidth}
                                >{\raggedright\arraybackslash}X@{}}
\toprule
\textbf{Label} & \textbf{Expected response} & \textbf{Definition} \\
\midrule
self-care &
No clinical response needed &
Manageable at home without clinical input; includes informational, retrospective, or low-risk questions without active concerning symptoms. \\
\addlinespace

schedule-visit &
Days to weeks &
Needs clinical assessment but is not urgent; includes persistent symptoms, medication/referral needs, or non-urgent follow-up questions. \\
\addlinespace

urgent-clinician-review &
Within 24--48 hours &
Requires timely clinical response due to worsening, moderate/severe, infectious, neurological, or special-population concerns that are not clearly emergent. \\
\addlinespace

emergency-referral &
Immediate &
Potentially life-threatening or time-critical presentation requiring immediate emergency care. \\
\bottomrule
\end{tabularx}
\caption{Definitions and expected response timeframes for the four actionable triage labels.}
\label{tab:appendix_label_definitions}
\end{table}

\subsection{Core decision principles}\label{apn:annotation-core-decision-principles}

\paragraph{Triage from text alone.}
Annotators assigned triage labels using only the patient inquiry text and the information available in the message. The task was not to infer a diagnosis. If clinical context was missing, annotators assigned the safest reasonable triage label and marked insufficient-info flag as true when appropriate.

\paragraph{Label by the most severe signal.}
When a message contained multiple symptoms or concerns, the label was assigned based on the most severe clinically relevant signal.

\paragraph{Prefer higher acuity when adjacent labels are ambiguous.}
When a case was genuinely ambiguous between two adjacent acuity levels, the higher-acuity label was selected. For example, if a case was borderline between \texttt{urgent\allowbreak-\allowbreak clinician\allowbreak-\allowbreak review} and \texttt{emergency\allowbreak-\allowbreak referral}, the guideline instructed annotators to choose \texttt{emergency\allowbreak-\allowbreak referral}.

\paragraph{Distinguish active symptoms from informational questions.}
Annotation began with a two-step distinction. First, annotators determined whether the patient, or another person described by the patient, was experiencing active symptoms and was not already under care. If yes, severity-based triage rules were applied. If no, the message was treated as informational. Second, for informational messages, annotators determined whether the inquiry asked for general information or an actionable next step. Table~\ref{tab:appendix_informational_classification} summarizes the typical label assignments for informational questions.

% \begin{table}[H]
% \centering
% \renewcommand{\arraystretch}{1.3}
% \resizebox{\columnwidth}{!}{%
% \begin{tabular}{>{\raggedright\arraybackslash}p{0.65\columnwidth}|>{\raggedright\arraybackslash}p{0.30\columnwidth}}
% \hline
% \textbf{Informational question type} & \textbf{Typical label} \\
% \midrule
% General knowledge or education & self-care \\
% Explanation of past symptoms or already evaluated findings & self-care \\
% Lifestyle or medication information without active risk & self-care \\
% Asking whether to seek care or what next step to take & schedule-visit \\
% Unresolved recurring condition requiring evaluation & schedule-visit \\
% \hline
% \end{tabular}%
% }
% \caption{Typical label assignments for informational patient inquiries.}
% \label{tab:appendix_informational_classification}
% \end{table}
\begin{table}[H]
\centering
\small
\renewcommand{\arraystretch}{1.15}
\begin{tabularx}{\columnwidth}{@{}>{\raggedright\arraybackslash}p{0.28\columnwidth} >{\raggedright\arraybackslash}X@{}}
\toprule
\textbf{Typical label} & \textbf{Informational question type} \\
\midrule
self-care & General knowledge or education \\
self-care & Explanation of past symptoms or already evaluated findings \\
self-care & Lifestyle or medication information without active risk \\
\addlinespace[0.4em]
\midrule
\addlinespace[0.2em]
schedule-visit & Asking whether to seek care or what next step to take \\
schedule-visit & Unresolved recurring condition requiring evaluation \\
\bottomrule
\end{tabularx}
\caption{Typical label assignments for informational patient inquiries.}
\label{tab:appendix_informational_classification}
\end{table}

\paragraph{Special population thresholds.}
The guideline used a lower threshold for escalation when the inquiry involved infants, elderly patients, pregnancy, immunocompromised status, or other vulnerability factors.

\paragraph{Known diagnosis or already under care.}
If symptoms were expected or stable for a known condition, and there was no new worsening signal, the case was usually treated as lower acuity. If the message asked for next-step action, follow-up, or unresolved management, \texttt{schedule\allowbreak-\allowbreak visit} could be appropriate.

\subsection{Detailed label criteria}
\label{apn:annotation-detailed-label-criteria}

\paragraph{\texttt{self\allowbreak-\allowbreak care}.}
The \texttt{self\allowbreak-\allowbreak care} label included mild cold symptoms or minor discomfort of short duration; general wellness, diet, exercise, or prevention questions; stable known chronic conditions without new concerning symptoms; curiosity or educational questions; retrospective questions about already resolved symptoms; and already evaluated cases asking for general explanation rather than action.

\paragraph{\texttt{schedule\allowbreak-\allowbreak visit}.}
The schedule-visit label included persistent symptoms, especially beyond approximately one week; medication adjustment questions; referral or specialist evaluation needs; mild new symptoms without red flags; non-urgent follow-up care; lab or imaging abnormalities without acute danger signs; and recurrent unexplained episodes without immediate danger signs.

\paragraph{\texttt{urgent\allowbreak-\allowbreak clinician\allowbreak-\allowbreak review}.}
The urgent-clinician-review label included worsening symptoms; moderate or severe pain without immediate emergency features; fever lasting more than 48 hours; infection signs such as pus, spreading redness, warmth, or swelling; new neurological symptoms; infant fever; passive suicidal ideation without active plan; and pregnancy with concerning symptoms.

\paragraph{\texttt{emergency\allowbreak-\allowbreak referral}.}
The emergency-referral label included chest pain with radiation; acute shortness of breath; loss of consciousness; stroke-like symptoms; severe bleeding; seizures; suicidal intent with plan; severe allergic reaction; infant under 3 months with fever; and any immediate life-threatening risk.

\subsection{Insufficient-information handling}\label{apn:annotation-insufficient-information}

The insufficient-info field was an auxiliary flag. Annotators used insufficient-info = true when the message lacked important clinical details. Even then, annotators still assigned one of the four final triage labels using the safest reasonable interpretation.

\subsection{Edge-case rules}\label{apn:annotation-edge-case-rules}

Annotators used the most severe reasonable interpretation when symptom severity was unclear. If adjacent acuity levels were both plausible, the higher-acuity label was selected. Table~\ref{tab:appendix_common_patterns} summarizes additional edge-case rules for mental health inquiries, inquiries about another person, and common patient-message patterns.

% \begin{table}[H]
% \centering
% \renewcommand{\arraystretch}{1.3}
% \resizebox{\columnwidth}{!}{%
% \begin{tabular}{>{\raggedright\arraybackslash}p{0.58\columnwidth}|>{\raggedright\arraybackslash}p{0.37\columnwidth}}
% \hline
% \textbf{Pattern} & \textbf{Typical label} \\
% \midrule
% Chest pain, unclear severity & urgent-clinician-review unless red flags suggest emergency \\
% Acute severe chest pain, radiation, shortness of breath, or collapse & emergency-referral \\
% Dizziness without red flags & schedule-visit \\
% Child fever without high-risk features & schedule-visit or urgent-clinician-review, depending on age and duration \\
% Headache without red flags & schedule-visit \\
% Sudden neurological deficit & emergency-referral \\
% Lab abnormalities only, no acute symptoms & schedule-visit \\
% Past symptoms only, already resolved & self-care \\
% \hline
% \end{tabular}%
% }
% \caption{Common triage patterns and typical label assignments.}
% \label{tab:appendix_common_patterns}
% \end{table}

\begin{table}[H]
\centering
\small
\renewcommand{\arraystretch}{1.15}
\begin{tabularx}{\columnwidth}{@{}>{\raggedright\arraybackslash}X
                                >{\raggedright\arraybackslash}p{0.34\columnwidth}@{}}
\toprule
\textbf{Scenario or pattern} & \textbf{Typical label} \\
\midrule

\multicolumn{2}{@{}l}{\textit{Mental health}} \\
\midrule
Mild anxiety without safety concerns &
self-care or schedule-visit, depending on duration and impairment \\
Worsening depression &
urgent-clinician-review \\
Passive suicidal ideation without plan &
urgent-clinician-review \\
Active suicidal intent, plan, or imminent risk &
emergency-referral \\

\addlinespace[0.4em]
\midrule
\multicolumn{2}{@{}l}{\textit{For someone else}} \\
\midrule
Someone collapsed and unresponsive &
emergency-referral \\
Friend or family member with active suicidal plan &
emergency-referral \\
Baby with very high fever or concerning symptoms &
urgent-clinician-review or emergency-referral, depending on age and severity \\
Friend or family member already in hospital; asks about prognosis &
self-care or informational \\

\addlinespace[0.4em]
\midrule
\multicolumn{2}{@{}l}{\textit{Common patterns}} \\
\midrule
Chest pain, unclear severity &
urgent-clinician-review unless red flags suggest emergency \\
Acute severe chest pain, radiation, shortness of breath, or collapse &
emergency-referral \\
Dizziness without red flags &
schedule-visit \\
Child fever without high-risk features &
schedule-visit or urgent-clinician-review, depending on age and duration \\
Headache without red flags &
schedule-visit \\
Sudden neurological deficit &
emergency-referral \\
Lab abnormalities only, no acute symptoms &
schedule-visit \\
Past symptoms only, already resolved &
self-care \\

\bottomrule
\end{tabularx}
\caption{Edge-case triage rules for mental health inquiries, inquiries about another person, and common patient-message patterns.}
\label{tab:appendix_edge_case_rules}
\label{tab:appendix_mental_health}
\label{tab:appendix_about_someone_else}
\label{tab:appendix_common_patterns}
\end{table}

\subsection{Annotation record format}
\label{apn:annotation-record-format}

Annotators recorded structured information for each reviewed sample (json formatted for readability):

\begin{quote}
\noindent \{ \\
\hspace*{1em} "id": 28286, \\
\hspace*{1em} "label": "emergency-referral", \\
\hspace*{1em} "confidence": "high", \\
\hspace*{1em} "insufficient\_info": false, \\
\hspace*{1em} "missing\_info": null, \\
\hspace*{1em} "reasoning": "Brief explanation of the key triage signal and label decision.", \\
\hspace*{1em} "alternative\_label": null, \\
\hspace*{1em} "notes": "Optional free-text notes for edge cases." \\
\}
\end{quote}

Example with insufficient information:

\begin{quote}
\noindent \{ \\
\hspace*{1em} "id": 12345, \\
\hspace*{1em} "label": "urgent\allowbreak-\allowbreak clinician\allowbreak-\allowbreak review", \\
\hspace*{1em} "confidence": "medium", \\
\hspace*{1em} "insufficient\_info": true, \\
\hspace*{1em} "missing\_info": "Need vital signs, symptom duration, severity, and relevant clinical history.", \\
\hspace*{1em} "reasoning": "The message describes potentially concerning symptoms but lacks enough detail to distinguish between non-urgent and urgent evaluation. A higher-acuity label is selected using the safest reasonable interpretation.", \\
\hspace*{1em} "alternative\_label": "schedule-visit", \\
\hspace*{1em} "notes": "Could be lower acuity if symptoms are mild, chronic, and already evaluated." \\
\}
\end{quote}

\subsection{Initial-labeling prompt template}
\label{apn:initial-labeling-prompt-template}

\paragraph{System prompt.}
\begin{quote}
Rules Summary: \\
1. Label by the most severe signal. \\
2. Do not diagnose. \\
3. When uncertain, choose the more severe label. \\
4. Lower threshold for vulnerable populations. \\
5. Respond with JSON only.
\end{quote}

\paragraph{User prompt template.}
\begin{quote}
\noindent \{ \\
\hspace*{1em} "label": "<one of: self-care, schedule-visit, urgent\allowbreak-\allowbreak clinician\allowbreak-\allowbreak review, emergency\allowbreak-\allowbreak referral>", \\
\hspace*{1em} "confidence": "<high, medium, or low>", \\
\hspace*{1em} "reasoning": "<1-2 sentences explaining the triage decision>", \\
\hspace*{1em} "insufficient\_info": <true or false>, \\
\hspace*{1em} "missing\_info": "<what information is missing, or null>" \\
\}
\end{quote}

\subsection{Guideline examples}
\label{apn:guideline-examples}

\paragraph{\texttt{self\allowbreak-\allowbreak care}.}
Patient inquiry: ``I have a mild runny nose for 2 days, no fever.''
\begin{quote}
\noindent \{ \\
\hspace*{1em} "label": "self-care", \\
\hspace*{1em} "confidence": "high", \\
\hspace*{1em} "reasoning": "Mild cold symptoms of short duration with no red flags. Safe to manage at home.", \\
\hspace*{1em} "insufficient\_info": false \\
\}
\end{quote}

\paragraph{\texttt{urgent\allowbreak-\allowbreak clinician\allowbreak-\allowbreak review}.}
Patient inquiry: ``My surgical wound is getting redder and there's some pus.''
\begin{quote}
\noindent \{ \\
\hspace*{1em} "label": "urgent\allowbreak-\allowbreak clinician\allowbreak-\allowbreak review", \\
\hspace*{1em} "confidence": "high", \\
\hspace*{1em} "reasoning": "Signs of wound infection require clinical attention within 24-48 hours.", \\
\hspace*{1em} "insufficient\_info": false \\
\}
\end{quote}

\begin{figure*}[t]
    \centering
    \small
    \obeylines\obeyspaces
    \begin{promptbox}{System Prompt}
\#\#\# Role\: You are a clinical workflow triage classifier for online patient inquiries.
\#\#\# Task: Classify the patient message into exactly one triage label. This is for workflow routing, not diagnosis or treatment recommendation.
Allowed labels:
1. self-care
2. schedule-visit
3. urgent-clinician-review
4. emergency-referral\\
\#\#\# Definitions::
[self-care]
Use for general health questions, mild short-duration symptoms, stable known conditions, medication/lifestyle questions, retrospective checks, or informational questions that do not require prompt clinician evaluation.
[schedule-visit]
Use when the message suggests the patient should have a non-urgent clinician visit or routine evaluation within days to weeks. This includes persistent but stable symptoms, mild new findings, medication adjustment questions, specialist referral needs, or unresolved recurring issues without urgent red flags.
[urgent-clinician-review]
Use when the message suggests clinician review is needed within 24–48 hours. This includes worsening symptoms, moderate or severe pain, fever lasting more than 48 hours, signs of infection, non-sudden neurological symptoms, concerning symptoms in infants/elderly/immunocompromised patients, or passive suicidal ideation without stated plan or intent.
[emergency-referral]
Use when the message suggests immediate emergency risk. This includes severe or crushing chest pain, chest pain with radiation, severe shortness of breath, loss of consciousness, stroke-like symptoms, seizure, severe bleeding, severe allergic reaction, sepsis-like deterioration, infant under 3 months with fever, or suicidal ideation with plan or intent. \\
\#\#\# Decision rules:
- Assign the label based on the most severe signal in the message.
- Do not diagnose. Classify only the urgency of response.
- If the message is informational or retrospective and no one has active unmanaged symptoms, use self-care unless the patient asks for next-step action.
- If symptoms are active, worsening, persistent, or functionally impairing, do not use self-care.
- If a special population is involved, such as an infant, elderly patient, pregnancy, or immunocompromised patient, use a lower threshold for escalation.
- If uncertain between two adjacent severity levels, choose the higher-risk label.
- If the message is too vague but contains some clinical signal, still assign the safest reasonable triage label and set insufficient\_info to true.
- If the message has no usable clinical content, assign schedule-visit and set insufficient\_info to true. \\
\#\#\# Output format:
Return valid JSON only. Do not include markdown, explanations, or extra text.
Schema:
{
  "label": "self-care | schedule-visit | urgent-clinician-review | emergency-referral",
  "confidence": "high | medium | low",
  "insufficient\_info": true | false
} 
Patient message:
\texttt{\{patient\_message\}} \\
\end{promptbox}
    \caption{Shared base prompt used for the 0-, 4-, 12-shot.}
    \label{prompt:system_prompt}
\end{figure*}

\begin{figure*}[p]
    \centering
    \small
    \obeylines
    \begin{promptbox}{4-shot Examples}
\#\#\# Patient message:
"First of all thanks for your gesture in providing this query facility. Kindly enlighten why we get cold and cough when ever weather changes from hot to cold, and also during winter temperature? What are the effective home remedies for persistent common cold and cough?"
Output:
{"label":"self-care"} \\
Patient message:
"I was given amoxicillin for an tooth infection , I took it for three days then stop taken it as the pain went away. Now I believe I have what appears to be a bladder infection(frequent and painful urination). will I be able to take the medicine again? I am a 68 year old male."
Output:
{"label":"schedule-visit"} \\
Patient message:
"The problem is that when I go to bed and lie down I start to have a tightness in my chest and become breathless if I get back up and sit up it improves last night it was several hours before I could return to bed. My health history is I have had three heart attacks and have a triple a"
Output:
{"label":"urgent-clinician-review"} \\
Patient message:
"My sister has hepatitis c and her liver enzymes are very high 3000. Her stomach is very hard due to an inguinal hernia that isnt repaired. She had a high fever last night. Shes had ultrasound and MRI, what other imaging studies and tests should she have?"
Output:
{"label":"emergency-referral"}
\end{promptbox}
    \caption{Four labeled examples appended to the shared base prompt for the 4-shot condition. We include one example per triage class.}
    \label{prompt:4-shot-examples}
\end{figure*}

\begin{figure*}[p]
    \centering
    \small
    \obeylines
    \begin{promptbox}{12-shot Additional 8 Examples}
\#\#\# Patient message:
"Hi I am Carl I am 66 years of age and and recently had a stroke or heart attack (doctors report is inconclusive) I was in a coma for one month and when i came out i could not walk this all happened on the 1st of March 2014 i had physical theraphy for whole month of April and thanks to Almighty God I am walking again with out the aid of any thing I am on lots of meds..Two friends have been telling me of Herbal Meds Can you help answer this for me please"
Output:
{"label":"self-care"} \\
Patient message:
"I have been on Microgestin 1/20 Fe BC for around a year now. This is the second month in a row I haven t had a period. I have this problem in the past and my doctor said its pretty normal but has changed my BC anyways. Both this month and last I ve had very, very slight spotting for less than a day and then nothing. Is it normal for my BC to pretty much rid me of my period like this?"
Output:
{"label":"self-care"} \\
Patient message:
"Hi- This is urgent. My niece does this thing- it has happened four times in a year now- She will be normally playing or sitting or eating, and all of a sudden her body turns all pale- her lips are completely white and she goes unresponsive- she is awake and her eyes are open but her breathing becomes shallow and her mom and I blow in her nose which makes her take deep breaths untill emergancy assistance arrives. Then after about 15 minutes she snaps out of it. Today her MRI results came back normal, her EEG was also normal. What is going on?? What other steps can we take to get to the bottem of this scary problem??"
Output:
{"label":"schedule-visit"} \\
Patient message:
"I did a lot of excessive moving and cleaning my house for two days, I now have a lot of pain in lower back area, which I think is muscle spasm, I have had pain for 5 days now, kept me up late nights for 2 nights, what can I take or do to get rid of pain in lower back extending to both sides of my hips. Thank you"
Output:
{"label":"schedule-visit"} \\
Patient message:
"My 6 year old son has been ill for over a week now with headaches, and lumps on the back of his head. Been to A\&E and was told it was swollen glands and was given calpol. A few days later he was still very poorly and lumps had got bigger. Went to my GP who said it is a scalp infection and gave antibiotics and antihistamine. Finished course of medicine and lumps are still present and sore! What do i do next!!?"
Output:
{"label":"urgent-clinician-review"} \\
Patient message:
"hi how can i change my attittude... i have bad attittude . i cant honest for money dealing ... my english very poor ...but i hope u can understand. i cnt save any thing my life i cant keep my promisses .. now i am facing a big problem for money my mind is always thinking about suicide"
Output:
{"label":"urgent-clinician-review"} \\
Patient message:
"Hi how are you? I am in a lot of pain. I have been having abdominal pain for about 2 weeks now. t is severe pain. Started out in my lower abdomen and then increased to the left side. Now it is all over (abdomen) and sometimes in my shoulder. It hurts to walk. I cant sleep properly because it hurts to lay down. When i walk fast or take a deep breath my insides hurt. It almost feels as though my insides are crushed. My stomach feels like its swollen. The pain is very very bad. Do you know what this can be? I have went to the doctor. The gave me an ultrasound and i am suppose to go Friday to profound on it."
Output:
{"label":"emergency-referral"} \\
Patient message:
"No aparent reason otherwise healty male. Non smoker non drinker passes out stops breathing. Heart attack and stroke have been ruled out. Keeping body on 34 degrees giving nitro. Medical induced a coma medical induced parlysis. Full life support? Your best medically idea would be?"
Output:
{"label":"emergency-referral"}
\end{promptbox}
    \caption{Eight additional labeled examples appended on top of the 4-shot to construct the 12-shot. Together with the four examples in Figure~\ref{prompt:4-shot-examples}, these provide three examples per triage class.}
    \label{prompt:12-shot-examples}
\end{figure*}

\section{Additional Prompted LLM Details}
\label{apn:llm-source}
\subsection{Model identifiers and serving setup}
\label{apn:llm-model-identifiers}

Prompted LLMs were evaluated without parameter updates. API-served models were accessed through their respective APIs, and open-source models were served locally through Ollama on a Google Colab T4 runtime. Table~\ref{tab:llm_model_identifiers} reports the model identifiers, serving methods, and sources used in the experiments.

\begin{table}[H]
\centering
\renewcommand{\arraystretch}{1.3}
\resizebox{\columnwidth}{!}{%
\begin{tabular}{>{\raggedright\arraybackslash}p{0.25\columnwidth}|>{\raggedright\arraybackslash}p{0.30\columnwidth}|>{\raggedright\arraybackslash}p{0.18\columnwidth}|>{\raggedright\arraybackslash}p{0.22\columnwidth}}
\hline
\textbf{Display name} & \textbf{Model identifier} & \textbf{Serving method} & \textbf{Source} \\
\midrule
GPT-4o-mini & gpt-4o-mini & OpenAI API & OpenAI (2024) \\
Claude Haiku 4.5 & claude-haiku-4-5-20251001 & Anthropic API & Anthropic (2025) \\
Llama 3.1 8B & llama3.1:8b & Ollama & Grattafiori et al. (2024) \\
Qwen3 8B & qwen3:8b & Ollama & Yang et al. (2025) \\
Mistral 7B & mistral:7b & Ollama & Jiang et al. (2023) \\
DeepSeek-R1 Distill 7B & deepseek-r1:7b & Ollama & DeepSeek-AI et al. (2025) \\
\hline
\end{tabular}%
}
\caption{Prompted LLM model identifiers, serving methods, and sources. Model identifiers correspond to the exact API or Ollama tags used in inference; sources identify the corresponding model family or official provider documentation.}
\label{tab:llm_model_identifiers}
\end{table}

\subsection{Inference settings}\label{apn:llm-inference-settings}

Table~\ref{tab:llm_inference_settings} summarizes the inference settings, and other generation options were left at defaults.

\begin{table}[H]
\centering
\renewcommand{\arraystretch}{1.3}
\resizebox{\columnwidth}{!}{%
\begin{tabular}{>{\raggedright\arraybackslash}p{0.55\columnwidth}|>{\raggedright\arraybackslash}p{0.40\columnwidth}}
\hline
\textbf{Setting} & \textbf{Value} \\
\midrule
Temperature & 0.0 \\
Calls per item & 1 \\
Self-consistency / majority vote & Not used \\
API max output tokens & 100 \\
Ollama stream mode & False \\
Ollama keep-alive & 60m \\
\hline
\end{tabular}%
}
\caption{LLM inference settings used across prompted model evaluations.}
\label{tab:llm_inference_settings}
\end{table}

\subsection{Shared prompt template and output schema} \label{apn:llm-shared-prompt-template}
All prompted LLMs used the same triage instruction template, which framed the task as workflow-oriented triage classification and required structured JSON output. We evaluated three prompting conditions: 0-shot, 4-shot, and 12-shot. The 0-shot condition used only the base prompt; the 4-shot condition added one labeled demonstration per class; and the 12-shot condition added three demonstrations per class, including the four examples used in the 4-shot prompt.

The appendix reports prompting materials incrementally: Figure~\ref{prompt:system_prompt} shows the shared base prompt, Figure~\ref{prompt:4-shot-examples} shows the four 4-shot demonstrations, and Figure~\ref{prompt:12-shot-examples} shows the eight additional demonstrations used to construct the 12-shot prompt.

\section{Additional Experiment Details}
\begin{table*}[ht]
\centering
\small
\setlength{\tabcolsep}{5pt}
\renewcommand{\arraystretch}{1.12}
\begin{tabular}{@{}p{0.47\textwidth}rrrr@{}}
\toprule
\textbf{Comparison (A vs. B)} 
& \makecell{\textbf{A right,}\\\textbf{B wrong} \\ \(\boldsymbol{b}\)}
& \makecell{\textbf{A wrong,}\\\textbf{B right} \\ \(\boldsymbol{c}\)}
& \makecell{\textbf{Two-sided}\\\textbf{exact } \(\boldsymbol{p}\)}
& \makecell{\textbf{Sig.}\\\(\boldsymbol{\alpha=0.05}\)} \\
\midrule

Claude Haiku 4.5 12-shot vs. Llama 3.1 8B 12-shot
& 38 & 48 & 0.332 & n.s. \\

Claude Haiku 4.5 12-shot vs. Qwen3-8B 4-shot
& 37 & 38 & 1.000 & n.s. \\

Llama 3.1 8B 12-shot vs. Qwen3-8B 4-shot
& 37 & 28 & 0.321 & n.s. \\

Claude Haiku 4.5 12-shot vs. GPT-4o-mini 12-shot
& 47 & 27 & 0.027 & \(^{*}\) \\

Llama 3.1 8B 12-shot vs. GPT-4o-mini 12-shot
& 55 & 25 & 0.001 & \(^{***}\) \\

Qwen3-8B 4-shot vs. GPT-4o-mini 12-shot
& 45 & 24 & 0.015 & \(^{*}\) \\

Claude Haiku 4.5 12-shot vs. BioBERT-v1.1 default
& 62 & 49 & 0.255 & n.s. \\

Llama 3.1 8B 12-shot vs. BioBERT-v1.1 default
& 70 & 47 & 0.042 & \(^{*}\) \\

Qwen3-8B 4-shot vs. BioBERT-v1.1 default
& 60 & 46 & 0.206 & n.s. \\

GPT-4o-mini 12-shot vs. BioBERT-v1.1 default
& 54 & 61 & 0.576 & n.s. \\
\bottomrule
\end{tabular}
\caption{Pairwise McNemar tests on per-example correctness. Tests use two-sided exact \(p\)-values on the gold set (\(n=300\)). Here, \(b\) denotes examples where model A is correct and model B is wrong, while \(c\) denotes examples where model A is wrong and model B is correct. Significance codes: \(^{*}p<0.05\), \(^{**}p<0.01\), \(^{***}p<0.001\).}
\label{tab:mcnemar_pairwise}
\end{table*}
\begin{table*}[ht]
\centering
\small
\setlength{\tabcolsep}{7pt}
\renewcommand{\arraystretch}{1.15}

\begin{tabular}{@{}clcccc@{}}
\toprule
\textbf{ID}
& \textbf{Model pair}
& \makecell{\textbf{Self-care}\\\(\boldsymbol{F_1}\)}
& \makecell{\textbf{Schedule-visit}\\\(\boldsymbol{F_1}\)}
& \makecell{\textbf{Urgent}\\\(\boldsymbol{F_1}\)}
& \makecell{\textbf{Emergency}\\\(\boldsymbol{F_1}\)} \\
\midrule

P1
& \makecell[l]{GPT-4o-mini 12-shot +\\ Llama 3.1 8B 12-shot}
& 0.869
& \textbf{0.729}
& \textbf{0.542}
& 0.692 \\

\addlinespace[4pt]

P2
& \makecell[l]{Claude Haiku 4.5 12-shot +\\ Llama 3.1 8B 12-shot}
& 0.845
& 0.720
& 0.523
& \textbf{0.717} \\

\addlinespace[4pt]

P3
& \makecell[l]{Llama 3.1 8B 12-shot +\\ Qwen3-8B 4-shot}
& \textbf{0.872}
& 0.714
& 0.522
& 0.655 \\

\addlinespace[4pt]

P4
& \makecell[l]{Claude Haiku 4.5 12-shot +\\ Qwen3-8B 4-shot}
& 0.824
& 0.697
& 0.464
& 0.633 \\

\addlinespace[4pt]

P5
& \makecell[l]{Llama 3.1 8B 12-shot +\\ Qwen3-8B 12-shot}
& 0.856
& 0.690
& 0.491
& 0.615 \\

\bottomrule
\end{tabular}
\caption{Per-class oracle-HITL \(F_1\) by two-model consensus pair.}
\label{tab:oracle_hitl_per_class}
\end{table*}
\begin{table*}[ht]
\centering
\small
\setlength{\tabcolsep}{7pt}
\renewcommand{\arraystretch}{1.15}

\begin{tabular}{@{}clcccc@{}}
\toprule
\textbf{ID}
& \textbf{Model pair}
& \makecell{\textbf{Self-care}\\\textbf{accuracy} \((n)\)}
& \makecell{\textbf{Schedule-visit}\\\textbf{accuracy} \((n)\)}
& \makecell{\textbf{Urgent-clinician-review}\\\textbf{accuracy} \((n)\)}
& \makecell{\textbf{Emergency-referral}\\\textbf{accuracy} \((n)\)} \\
\midrule

P1
& \makecell[l]{GPT-4o-mini 12-shot +\\ Llama 3.1 8B 12-shot}
& 94.3\% (53)
& 57.1\% (35)
& 25.8\% (66)
& \textbf{55.6\%} (18) \\

\addlinespace[4pt]

P2
& \makecell[l]{Claude Haiku 4.5 12-shot +\\ Llama 3.1 8B 12-shot}
& \textbf{97.9\%} (47)
& 61.1\% (61)
& 26.3\% (57)
& \textbf{55.6\%} (18) \\

\addlinespace[4pt]

P3
& \makecell[l]{Llama 3.1 8B 12-shot +\\ Qwen3-8B 4-shot}
& 91.7\% (72)
& \textbf{63.4\%} (41)
& 23.7\% (59)
& 45.0\% (20) \\

\addlinespace[4pt]

P4
& \makecell[l]{Claude Haiku 4.5 12-shot +\\ Qwen3-8B 4-shot}
& 94.2\% (52)
& 54.0\% (50)
& 24.6\% (65)
& 46.4\% (28) \\

\addlinespace[4pt]

P5
& \makecell[l]{Llama 3.1 8B 12-shot +\\ Qwen3-8B 12-shot}
& 90.7\% (75)
& 55.3\% (47)
& \textbf{27.0\%} (63)
& 47.4\% (19) \\

\bottomrule
\end{tabular}
\caption{Per-class consensus accuracy by two-model pair. Auto-accepted cases are those where the two models agree; disagreement cases are excluded and treated as manual escalations. For each predicted label, \(n\) denotes the number of auto-accepted cases assigned to that label.}
\label{tab:per_class_consensus_accuracy}

\end{table*}
\subsection{Pairwise McNemar test results} \label{apn:mcnemar}

Table \ref{tab:mcnemar_pairwise} reports pairwise McNemar tests on per-example correctness for selected models on the 300-example gold set. Most comparisons were not statistically significant, indicating that several top-performing LLM settings had similar paired accuracy. Significant differences were observed for comparisons involving GPT-4o-mini 12-shot versus Claude Haiku 4.5 12-shot, Llama 3.1 8B 12-shot, and Qwen3-8B 4-shot, as well as Llama 3.1 8B 12-shot versus BioBERT-v1.1 default. These paired tests complement macro-\(F_1\) and safety-oriented metrics by assessing whether two models differ in example-level correctness.

\subsection{Per-Class Consensus and Oracle-HITL Results} \label{apn:HITL}
Tables~\ref{tab:oracle_hitl_per_class} and~\ref{tab:per_class_consensus_accuracy} summarize per-class behavior in the two-model consensus workflow. Oracle-HITL \(F_1\) reflects full-cohort performance when disagreement cases are resolved correctly by human review, whereas consensus accuracy is computed only on auto-accepted cases where both models agree. The results suggest that consensus-based HITL improves per-class performance overall, but agreement is not equally reliable across classes, especially for urgent-clinician-review predictions.

\end{document}